\documentclass[final]{cvpr}
\usepackage{times}
\usepackage{epsfig}
\usepackage{graphicx}
\usepackage{amsmath}
\usepackage{amssymb}
\usepackage{braket}
\usepackage{mathtools}
\DeclarePairedDelimiter\abs{\lvert}{\rvert}%
\DeclarePairedDelimiter\norm{\lVert}{\rVert}%

\makeatletter
\let\oldabs\abs
\def\abs{\@ifstar{\oldabs}{\oldabs*}}
\let\oldnorm\norm
\def\norm{\@ifstar{\oldnorm}{\oldnorm*}}
\makeatother

\usepackage[pagebackref=true,breaklinks=true,colorlinks,bookmarks=false]{hyperref}

\def\cvprPaperID{8179} % *** Enter the CVPR Paper ID here
\def\confYear{CVPR 2021}

\newcommand\blfootnote[1]{%
  \begingroup
  \renewcommand\thefootnote{}\footnote{#1}%
  \addtocounter{footnote}{-1}%
  \endgroup
}

\begin{document}

%%%%%%%%% TITLE
\title{VDSM: Unsupervised Video Disentanglement with State-Space Modeling and Deep Mixtures of Experts}

\author{Matthew J. Vowels\\
{\tt\small m.j.vowels@surrey.ac.uk}
\and
Necati Cihan Camgoz\\
{\tt\small n.camgoz@surrey.ac.uk}

\and
Richard Bowden\\
{\tt\small r.bowden@surrey.ac.uk}
\and
Centre for Vision, Speech and Signal Processing \\
University of Surrey\\
Guildford, UK\\
}

\maketitle

\begin{abstract}
Disentangled representations support a range of downstream tasks including causal reasoning, generative modeling, and fair machine learning. Unfortunately, disentanglement has been shown to be impossible without the incorporation of supervision or inductive bias. Given that supervision is often expensive or infeasible to acquire, we choose to incorporate structural inductive bias and present an unsupervised, deep State-Space-Model for Video Disentanglement (VDSM). The model disentangles latent time-varying and dynamic factors via the incorporation of hierarchical structure with a dynamic prior and a Mixture of Experts decoder. VDSM learns separate disentangled representations for the identity of the object or person in the video, and for the action being performed. We evaluate VDSM across a range of qualitative and quantitative tasks including identity and dynamics transfer, sequence generation, Fr\'echet Inception Distance, and factor classification. VDSM achieves state-of-the-art performance and exceeds adversarial methods, even when the methods use additional supervision. 
\end{abstract}
\blfootnote{Accepted to CVPR 2021}

%%%%%%%%% BODY TEXT
\section{Introduction}
In general, humans are able to reason about the identity of an object and the object's motion independently, thereby implying that identity and motion are considered as disentangled generative attributes \cite{Higgins2018, bengio1}. In other words, a change to an object's motion does not affect the object's identity. For example, in sign language translation, the canonical form of a gesture exists independently of the identity or appearance of the signer. In order to reason independently about the latent factors underlying identity and motion, it is therefore desirable to seek disentanglement.

Achieving disentangled representations is a long standing goal for machine learning, and supports causal reasoning \cite{Parascandolo2018,Besserve2019, suter2018, Vowels2020b}, fair machine learning \cite{locatello2019fairness, louizos2017, creager2019, Vowels2020}, generalizability \cite{hosoya2020, MLVAE}, structured/controllable inference and prediction \cite{Yan2019}, attribute transfer \cite{bousmalis2016,press1}, and improved performance on downstream tasks \cite{bengio1, Lake2017, Ridgeway2016, Raffin2019, gatedvae}. Unfortunately, being able to consistently learn a disentangled representation such that the factors correspond with meaningful attributes has been shown to be both theoretically and empirically impossible without the use of some form of supervision or inductive bias \cite{dai3, locatello}. However, acquiring high-quality supervision is expensive and time-consuming. Whilst many methods rely on such supervision, we consider how the implicit structure embedded in video data can be leveraged and reflected in the structure of the model, in order to achieve unsupervised disentanglement.

\begin{figure}
\centering
\includegraphics[width=1\linewidth]{ 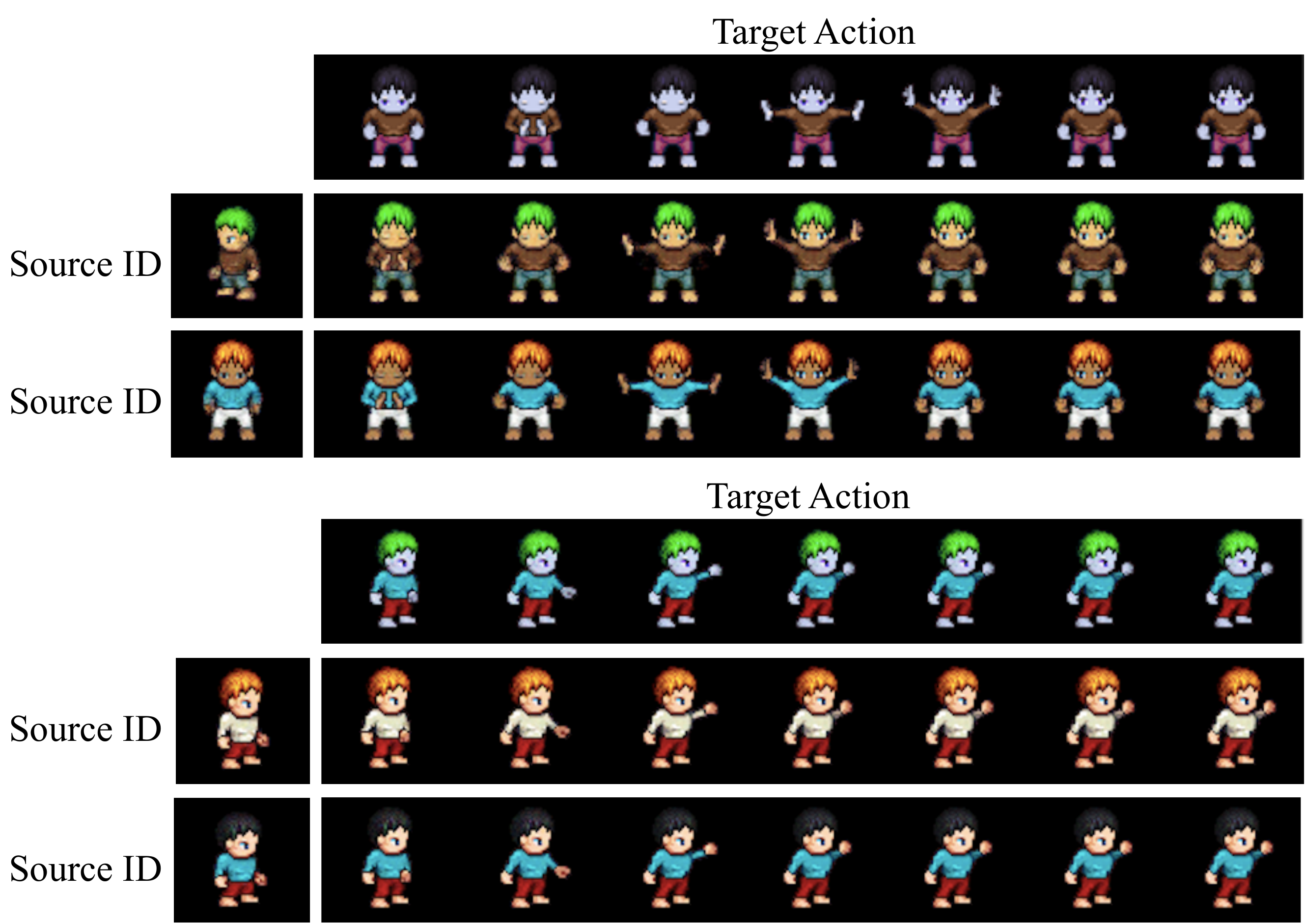}
\caption{VDSM action transfer for the Sprites dataset \cite{Li2018b}. The target action embedding is transferred swapped with one from a different identity and the sequence is generated.}
\label{fig:sprites_action_swap}
\end{figure}

We propose Video Disentanglement via State-Space-Modeling (VDSM). VDSM is motivated by a careful consideration of the generative structure of a video sequence, which is assumed to be composed of identity (i.e. the canonical appearance of an object or person), an action (i.e. dictating the dynamics governing change over time), and pose (i.e. the time varying aspects of appearance). 
%VDSM achieves unsupervised sequence generation and factor disentanglement through a novel, latent hierarchy with a nested sequence-to-sequence network and a deep mixture of experts decoder. By carefully constraining the structure of our learning process and the network structure, VDSM provides separate and independent embeddings for action and identity. VDSM is flexible in that it can also be used on groups of non-sequential images to disentangle identity from pose. 
In summary, VDSM: 
1) Is a completely unsupervised approach that avoids the need for adversarial training.
2) Incorporates a novel structure designed to factorise appearance and motion, using a strong mixture of decoders to separate identity. 
3) Produces embeddings that achieve state-of-the-art classification performance.
4) Far exceeds GAN based approaches in sequence generation (evidenced by FID scores).
5) Produces superior disentanglement compared to approaches which use forms of supervision. 
6) Exceeds accuracy consistency by over 30\% compared to the nearest competing approach. %(on the mug dataset *Table 2) 
7) Produces qualitative results that reflect the quantitative performance in terms of disentanglement and image quality.

The rest of this paper is structured as follows. First we discuss related work in Section 2 before describing the structure and training of VDSM in Section 3. We provide qualitative and quantitative experiments in Section 4 and conclude in Section 5.

\section{Related Work}

Disentanglement is an ongoing challenge for machine learning. However, achieving consistent, interpretable disentanglement without some form of supervision or inductive bias is impossible \cite{locatello}. A large body of recent work seeks disentanglement via the incorporation of various levels of weak- \cite{gatedvae, locatello2020, denton2017, MLVAE, shu2019, chen2019} or semi-supervision \cite{louizos2017, moyer1, DIVA, siddharth2017}. Many of these methods seek disentanglement between static and time-varying factors in sequences (such as content and pose). Impressive results have been achieved in extending this goal to multi-object tracking \cite{Akhundov2019,Kosiorek2018,  Eslami2016, Kossen2019,Stelzner2019, Hsieh2018}. Such an approach has the advantage of trading the need for explicit supervision with structural inductive bias. For example, Grathwohl \& Wilson (2016) \cite{Grathwohl2016} utilize variational inference with hierarchical latent factors and a slow-feature regularizing penalty to disentangle content and pose. Denton \& Birodkar (2017) \cite{denton2017} disentangle content from pose by combining adversarial training with two deterministic, recurrent encoder-decoders. They incorporate an adversarial component which discriminates between content encodings from within the same or between different sequences. Other methods which seek disentanglement between static and time-varying factors include: S3VAE \cite{Zhu2020}, which uses recurrent variational autoencoders; VideoVAE \cite{He2018}, a semi-supervised technique that enables attribute control for video generation; Factorizing VAE (FAVAE) \cite{Yamada2020}, which uses multi-scale time-based convolution with variational autoencoders; Disentangled Sequental Autoencoders (DSA) \cite{Li2018b} and Disentangled State Space Model (DSSM) \cite{Miladinovic2019} which undertake structured variational inference; and  G3AN \cite{Wang2020b} and MoCoGAN \cite{Tulyakov2017} both Generative Adversarial Networks (GANs).

Other networks which leverage structure to model sequence dynamics include a wide range of latent variable and state-space variational methods. For instance, Deep Kalman Filters \cite{Krishnan2015} extend the structure of the traditional Kalman filter to incorporate recent deep learning techniques for video prediction. Structured inference networks \cite{Krishnan2016} leverage similar structural considerations for sequential data, facilitating time-series causal inference from the disentangled representations. These networks are similar to other stochastic latent variable networks for sequence modeling, which tend to vary according to the structure imposed in the generative and inference models (see \textit{e.g.} \cite{Becker2019, YinND, Watter2015, Hafner2019b, Hafner2019, Rybkin2019, Schmeckpeper2019, Bayer2015, Chung2016, Marino2019, Aksan2019, Watter2015, Gregor2019, Fraccaro2016, Goyal2017}).

Not all methods are concerned with achieving factor disentanglement, and many are primarily designed to generate or predict future video frames. These networks may still include structured variational and/or adversarial techniques. Recent attempts at video generation include SinGAN \cite{Shaham2019} which produces a video sequence from a single image; various multi-scale generative adversarial networks \cite{Mathieu2016b, Clark2019, Acharya2018}; the Latent Video Transformer \cite{Rakhimov2020} which applies transformers \cite{Vaswani2017} to discrete latent representations learned in an autoencoder \cite{VQVAEs}; and pixel-level autoregressive methods \cite{Weissenborn2019, Kalchbrenner2017}. 

Evidently, the tasks of video generation and disentanglement overlap, particularly when structured networks are used. The more inductive bias is used, the more domain-specific the network tends to become. For example, the network may incorporate inductive bias corresponding to the physical laws of interaction and motion \cite{Fraccaro2017, Battaglia2016, Ye2018, Chen2019c, Toth2019, Greydanus2019}. We prefer to keep our method general, such that it may be applied to non-vision related tasks for which the benefits of such inductive bias may be inappropriate. 

 In terms of structure, our method `VDSM', is probably closest to DSA \cite{Li2018b}, Structured Inference Networks \cite{Krishnan2016}, DSSM \cite{Miladinovic2019} and Factorized Hierarchical VAE (FSVAE) \cite{Hsu2017}. In terms of the computer vision applications, VDSM is most similar to the recently released S3VAE \cite{Zhu2020} and G3AN \cite{Wang2020b}. The latter two methods seek to disentangle motion and appearance for video data, and incorporate various forms of weak- or semi-supervision, such as optical flow \cite{Zhu2020}, or labels for conditional motion generation \cite{Wang2020b}. Importantly, and in contrast with these two methods, our network does not use any form of supervision.

 Whilst adversarial methods such as G3AN and MoCoGAN are popular and have been shown to work well for density estimation and sequence generation, they are also notoriously difficult and unreliable to train \cite{moyer1, lezama, gabbay2019}. Furthermore, recent work has highlighted that adversarial training is potentially unnecessary, and that non-adversarial methods can achieve comparable or better results across a wide range of tasks \cite{moyer1, VQVAEs, Ho2020, Vahdat2020NVAE, Xiao2020VAEBM, gabbay2019}. Finally, supervision may not always be available, and unsupervised methods are more generally applicable in scenarios where labels (even partial labels) are not available. VDSM intends to address the considerable challenge of disentangling key generative factors in sequences (pose, identity, dynamics) without supervision, and without adversarial training.

 \begin{figure}
\centering
\includegraphics[width=.8\linewidth]{ 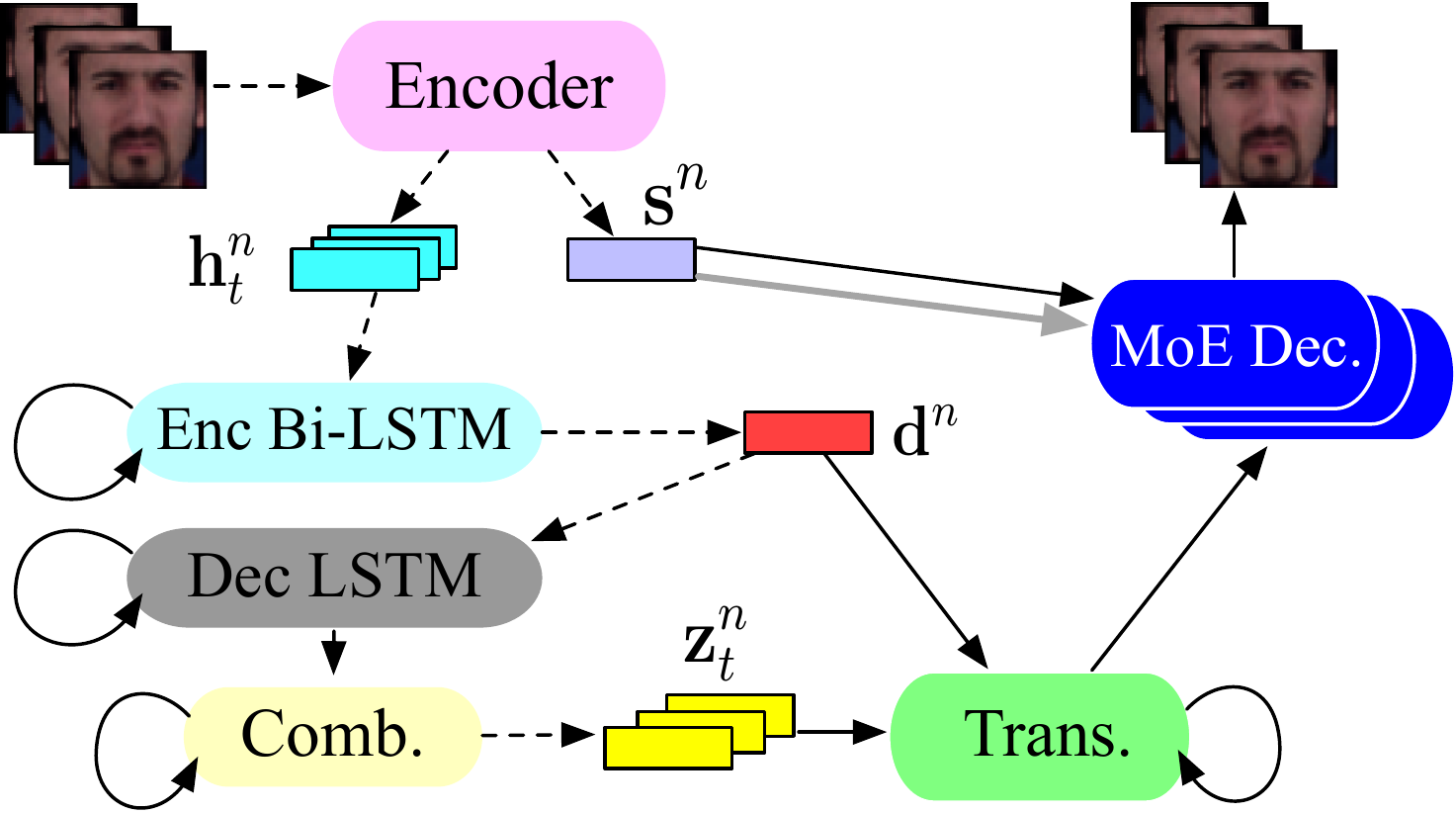}
\caption{Block diagram of VDSM architecture. Gray lines indicate mixture of expert selection, dashed lines indicate inference, solid lines indicate generation, and looped arrows indicate autoregressive dependency.}
\label{fig:block}
\end{figure}

 \begin{figure}
\centering
\includegraphics[width=.85\linewidth]{ 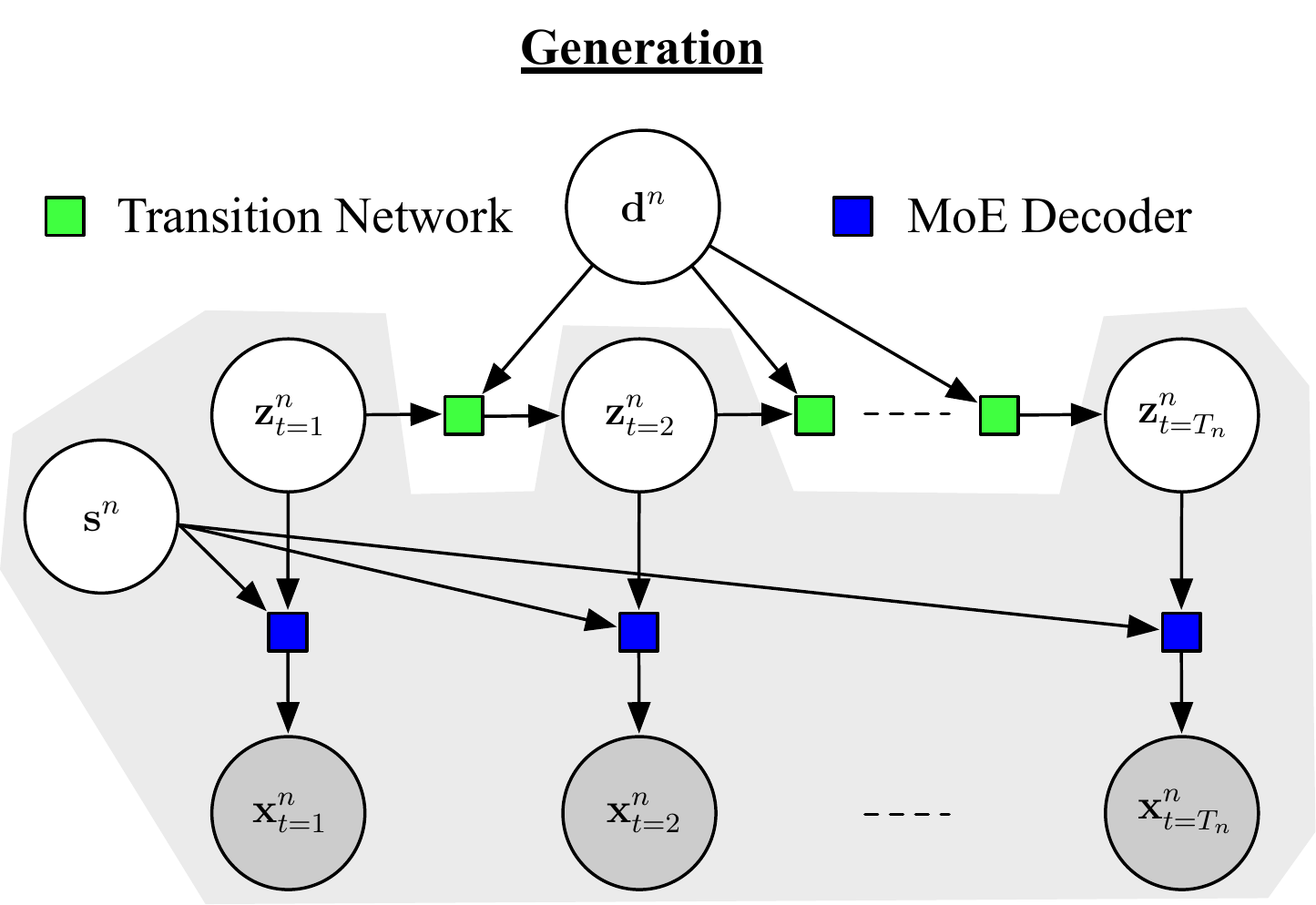}
\caption{Generative model for VDSM with images $\mathbf{x}$, latent pose $\mathbf{z}$, motion dynamics $\mathbf{d}$, and static factors / content $\mathbf{s}$. Grey shading indicates pre-trained components.}
\label{fig:genmodel}
\end{figure}

\section{Video Disentanglement with State-Space Modeling (VDSM)}
%\vspace{-8mm}
Consider $N$ sequences of natural images $\{\mathbf{x}_{1:T_n}^n \}_{n=1}^N$ for which we may assume each image in a sequence lies on some significantly lower-dimensional manifold. We wish to encode the images via a stochastic mapping $p_\theta(\mathbf{z}_t^n, \mathbf{s}^n, \mathbf{d}^n | \mathbf{x}_{t=1:T_n}^n)$ where $\mathbf{x}_{t=1:T_n}^n$ is the $n^{th}$ sequence of $T_n$ images, and $\mathbf{z}  \in \mathbb{R}^{\kappa_z}$,  $\mathbf{s} \in \mathbb{R}^{\kappa_s}$ and $\mathbf{d} \in \mathbb{R}^{\kappa_d}$ provide compact representations for the time varying latent factors (\textit{e.g.} pose), static latent factors (\textit{e.g.} identity), and the action dynamics (\textit{e.g.} waving), respectively. 

 A block diagram for VDSM is shown in Figure \ref{fig:block}. Inspecting this figure, it can be seen that images from a sequence are first encoded individually into two latent partitions $\mathbf{h}_t^n$ (which vary over the sequence), and $\mathbf{s}^n$ (which remain constant throughout a sequence. The time varying features are summarized by $\mathbf{d}^n$ which represents the dynamics or action being performed, and then decoded following a sequence-to-sequence (seq2seq) structure \cite{sutskever2}. A combination network (in yellow) takes the output of the seq2seq network and parameterizes a posterior distribution $\mathbf{z}_t^n$ which, when combined with the dynamics $\mathbf{d}^n$ and the static factors $\mathbf{s}^n$, are decoded to reconstruct the original images. In the decoding stage, the static factors $\mathbf{s}^n$ are also used to blend weights from a bank of decoders that form the Mixture of Experts (MoE) decoder which specializes in reconstructing information relevant to identity.
 
 According to the task of image reconstruction through a structured, probabilistic bottleneck, VDSM resembles a variational autoencoder \cite{kingma, rezende2}. We incorporate inductive bias in the form of hierarchical latent structure, dynamic priors, Markov factorization, and architectural constraints (\textit{e.g.} MoE and seq2seq). The \textbf{generative model} for VDSM is shown in Figure \ref{fig:genmodel} and can be factorized as follows:
\vspace{-2mm}
\begin{equation}
\begin{split}
    p_{\theta}(\mathbf{x}^n_{t=1:T_n}, \mathbf{s}^n, \mathbf{d}^n, \mathbf{z}^n_{t=1:T_n}) = \\ p_{\theta}(\mathbf{s}^n)p_{\theta}(\mathbf{d}^n)p_{\theta}(\mathbf{z}_{t=1}^n)\prod_{t=2}^{T_n} p_{\theta}(\mathbf{x}^n_t|\mathbf{s}^n, \mathbf{z}_t^n) p_{\theta}(\mathbf{z}_{t}^n|\mathbf{z}_{t-1}^n, \mathbf{d}^n)\\
    p_{\theta}(\mathbf{s}^n) = \mathcal{N}(0, \frac{1}{\mathrm{N}_s})\\
    \hat{\mathbf{s}}^n = \mbox{softmax}\left(\mathbf{s}^n / \tau_{s}  \right) \\
    p_{\theta}(\mathbf{d}^n) = \mathcal{N} \left( \mathbf{0}, \mathbf{1} \right)  \\
    p_{\theta}(\mathbf{x}^n_t|\mathbf{s}^n, \mathbf{z}_t^n) = \mbox{Bern}\left(\mbox{DEC}_{MoE}(\hat{\mathbf{s}}^n, \{\mathbf{s}^n, \mathbf{z}_t^n\} \right) )\\
    p_{\theta}(\mathbf{z}^n_{t=1}) = \mathcal{N} \left( \mathbf{0}, \mathbf{1} \right)\\
     p_{\theta}(\mathbf{z}^n_{t>1}) = \mathcal{N} \left( \mu_{tr}(\mathbf{z}_{t-1}^n, \mathbf{d}^n), \sigma_{tr}(\mathbf{z}_{t-1}^n, \mathbf{d}^n) \right)
    \end{split}
    \label{eq:generativeparams}
\end{equation}

Considering this factorization and Figure \ref{fig:genmodel}, for sequence $n$ we sample an action from the dynamics/action factor $\mathbf{d}^n$ and an initial pose $\mathbf{z}^n_{t=1}$. These are fed into a transition network (green) which produces the next pose factor $\mathbf{z}^n_{t=2}$ and we repeat this for $T_n$ timepoints. The transition network outputs the location $\mu_{tr}(\mathbf{z}_{t-1}^n, \mathbf{d}^n)$ and scale  $\sigma_{tr}(\mathbf{z}_{t-1}^n, \mathbf{d}^n)$ for a diagonal Gaussian parameterizing the next timepoint $\mathbf{z}_t^n$. We then sample an identity/static factor $\mathbf{s}^n$. For each timepoint, we concatenate this factor with the pose $\mathbf{z}_t$ for timepoints $t=1:T_n$ and pass them into a Mixture of Experts (MoE) Decoder, resulting in the conditional likelihood $p_{\theta}(\mathbf{x}^n_t|\mathbf{s}^n, \mathbf{z}_t^n)$. The MoE weights are determined by $\hat{\mathbf{s}}^n$ which is derived by dividing factors $\mathbf{s}^n$ by temperature $\tau_s$ and applying a softmax operation. The conditional likelihood (i.e., the distribution of generated images) is Bernoulli distributed according to Eq. \ref{eq:generativeparams} but can be chosen according to the data. $\mathrm{N}_s$ (line 3 Eq. \ref{eq:generativeparams}) represents a prior `guess' as to the number of distinct individuals/identities. $\hat{\mathbf{s}}^n$ is simplectic and similar to the Dirichlet distribution; it sums to 1, and the parameter $\tau_{s}$ controls the temperature \cite{Gelman1996}.\footnote{Similar to the Gumbel Softmax, or Concrete distributions \cite{Maddison2017, jang2017}.} A Dirichlet distribution was found to be less stable to train.

\begin{figure}
\centering
\includegraphics[width=.95\linewidth]{ 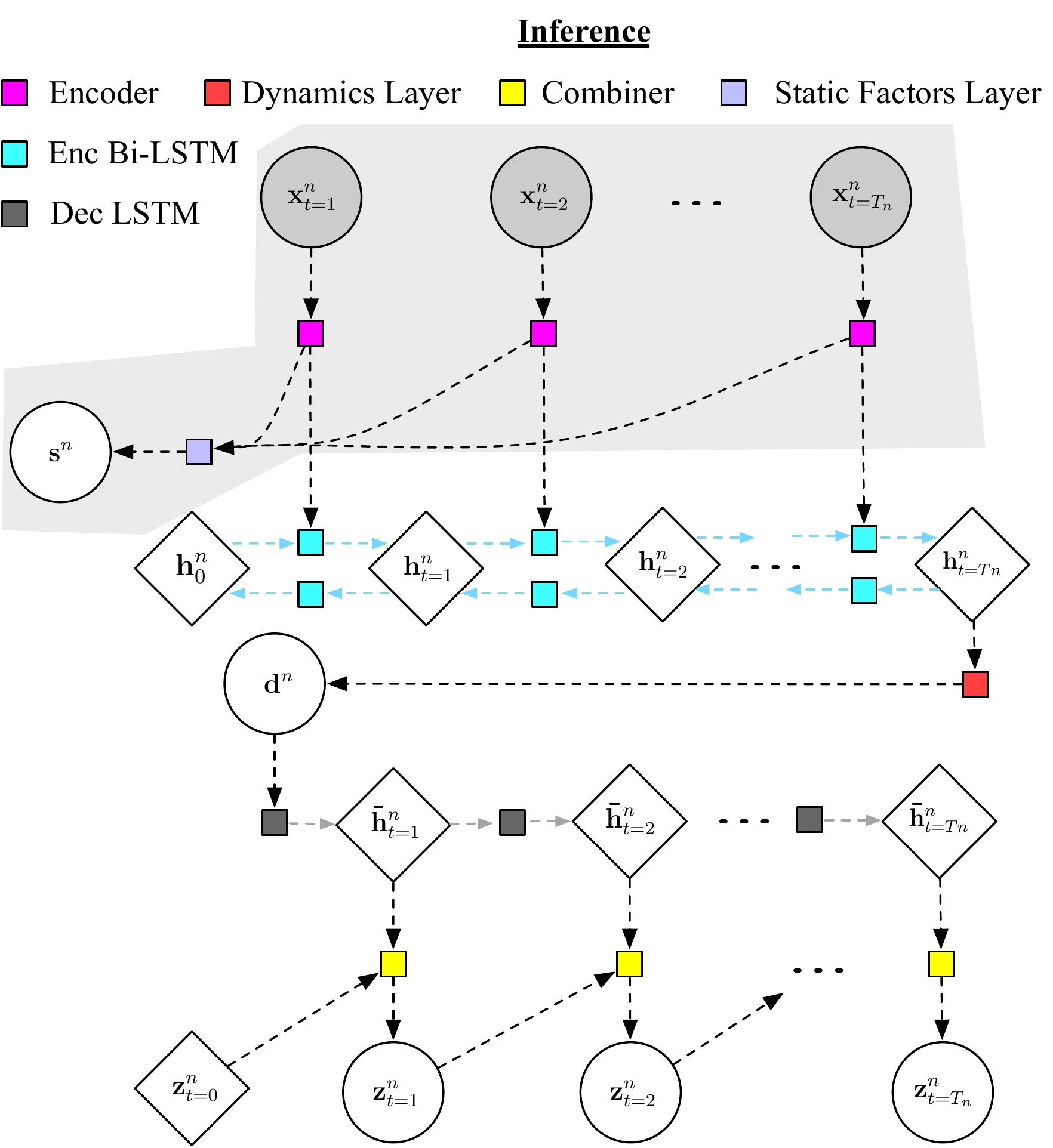}
\caption{Inference model for VDSM with images $\mathbf{x}$, latent pose $\mathbf{z}$, motion dynamics $\mathbf{d}$, encoder bi-LSTM and decoder LSTM hidden states $\mathbf{h}$ and $\bar{\mathbf{h}}$ resp., and static factors / content $\mathbf{s}$. \cite{Kingma2016}. Together the encoder and decoder LSTMs form a seq2seq model with dynamics $\mathbf{d}^n$ as the bottleneck/summary vector. Grey background shading indicates pre-trained components.}
\label{fig:infmodel}
\end{figure}

In order to undertake this inference, we leverage stochastic variational inference (SVI) \cite{Jordan1999, rezende2, kingma} to learn an approximate posterior distribution $q_{\phi}(.)$ according to the VDSM \textbf{inference model} shown in Equation \ref{eq:inf2}. In words, the static factor representing the identity of an individual $\mathbf{s}^n$ is represented as a diagonal-Gaussian distribution. The location and scale for this Gaussian are parameterized as functions $f_s$ of the average of the identity encodings of the images in a sequence using encoder $\mbox{ENC}_s(\mathbf{x}^n_{t=1:T_n})$. The function $f_s$ is a linear/fully-connected layer with a non-linear activation. By taking the average, we force the network to learn factors which remain constant over the course of a sequence (i.e., the identity). Before the softmax operation, $\tau_s$ is used to anneal the temperature of this distribution during training. The time-varying pose embeddings from the encoder $\mbox{ENC}_z$ (not shown in Eq. \ref{eq:inf2}) are fed into a recurrent, bidirectional Long Short Term Memory (bi-LSTM) \cite{hochreiterLSTM} network, the final hidden state of which $\mathbf{h}^n$ is fed through functions $f_{d\mu}(.)$ and $f_{d\sigma}(.)$ (also linear layers with non-linear activations) to parameterize a Gaussian distribution for dynamics factors $\mathbf{d}^n$. The dynamics embedding $\mathbf{d}^n$ is duplicated and used as the initial hidden and cell states for a decoding (uni-directional) LSTM. This decoder LSTM produces hidden states at each timestep $\bar{\mathbf{h}}_t$ which are fed to the combiner functions $\mu_{comb}$ and $\sigma_{comb}$, with $\mathbf{z}_{t-1}^n$ from the previous timestep.

\vspace{-3mm}
\begin{equation}
\begin{split}
    q_{\phi}(\mathbf{s}^n, \mathbf{d}^n, \mathbf{z}_{t=1:T_n} | \mathbf{x}^n_{t=1:T_n}) =\\ q_{\phi}(\mathbf{s}^n|\mathbf{x}^n_{t=1:T_n})q_{\phi}( \mathbf{d}^n|\mathbf{x}^n_{t=1:T_n})q_{\phi}(\mathbf{z}^n_{t=1},|\mathbf{x}^n_{t=1:T_n}, \mathbf{d}^n)\\ \prod_{t=2}^{T_n}q_{\phi}(\mathbf{z}^n_t| \mathbf{z}^n_{t-1},\mathbf{x}^n_{t=1:T_n}, \mathbf{d}^n)\\
        q_{\phi}(\mathbf{s}^n|\mathbf{x}^n_{t=1:T_n}) =  \mathcal{N} \left( s_{loc}, s_{scale} \right)\\
         s_{loc} = f_{s,loc}\left(\frac{1}{T_n}\sum_{t=1}^{T_n} (\mbox{ENC}_s(\mathbf{x}^n_{t}))\right) \\
          s_{scale} =f_{s,scale}\left(\frac{1}{T_n}\sum_{t=1}^{T_n}  (\mbox{ENC}_s(\mathbf{x}^n_{t}))\right)\\
        q_{\phi}(\mathbf{d}^n | \mathbf{x}^n_{t=1:T_n}) =  \left(\mathcal{N} f_{d\mu}(\mathbf{h}^n), f_{d\sigma}(\mathbf{n}^n) \right)  \\
        q_{\phi}(\mathbf{z}^n_{t} | \mathbf{x}^n_{t=1:T_n}, \mathbf{d}^n) =\\ \mathcal{N}(\mu_{comb}(\mathbf{z}_{t-1}^n,\bar{\mathbf{h}}_t^n, \mathbf{d}^n), \sigma_{comb}(\mathbf{z}_{t-1}^n,\bar{\mathbf{h}}_t, \mathbf{d}^n))) 
    \end{split}
    \label{eq:inf2}
\end{equation}
\vspace{-2mm} 
\begin{equation}
    \mbox{max}_{\phi, \theta} \Braket{ \Braket{ \frac{p_{\theta}(\mathbf{x}^n_{t=1:T_n}, \mathbf{s}^n, \mathbf{d}^n, \mathbf{z}^n_{t=1:T_n})}{q_{\phi}(\mathbf{s}^n, \mathbf{d}^n, \mathbf{z}_{t=1:T_n}^n\mid \mathbf{x}^n_{i=1:T_n})) }}_{q_{\phi}}}_{p_{\mathcal{D}(\mathbf{x}^n)}}
    \label{eq:elbo}
\end{equation}
The goal of the inference model is to make the problem of latent inference possible through the use of parametric approximating posteriors. Together, the generation and inference models can be optimized as part of a stochastic, amortized, variational inference objective, known as the Evidence Lower Bound (ELBO), given in Equation \ref{eq:elbo}. The ELBO objective may be derived similarly to \cite{Krishnan2016} (also see supplementary material) and is shown in Equation \ref{eq:fullobj}. In Eq. \ref{eq:fullobj}, each $\lambda$ term represents a hyperparameter used for annealing the corresponding KL (Kullback-Liebler divergence) objective term during training. 

\vspace{-2mm} 
\begin{equation}
\begin{split}
\mathcal{L}(\mathbf{x}^n_{1:T_n};(\theta,\phi)) =\\ \sum_{t=1}^{T_n} \mathbb{E}_{q_{\phi}(\mathbf{z}_t^n, \mathbf{s}^n | \mathbf{x}^n_{1:T_n})}[\log p_{\theta}(\mathbf{x}^n_t|\mathbf{z}_t^n, \mathbf{s}^n)]\\
-\lambda_{d}(\mbox{KL}(q_{\phi}(\mathbf{d}^n|\mathbf{x}^n_{1:T_n})||p_{\theta}(\mathbf{d}^n))) \\
-\lambda_{s}(\mbox{KL}(q_{\phi}(\mathbf{s}^n|\mathbf{x}^n_{1:T_n})||p_{\theta}(\mathbf{s}^n))) \\
-\lambda_{z}(\mbox{KL}(q_{\phi}(\mathbf{z}^n_1|\mathbf{x}^n_{1:T_n}, \mathbf{d}^n)||p_{\theta}(\mathbf{z}^n_1)) \\
- \lambda_{z} \sum_{t=2}^{T_n}\mathbb{E}_{q_{\phi}(\mathbf{z}_{t-1}^n|\mathbf{x}^n_{1:T_n},\mathbf{d}^n)}\\\mbox{KL}(q_{\phi}(\mathbf{z}_t^n|\mathbf{z}_{t-1}^n,\mathbf{d}^n, \mathbf{x}^n_{1:T_n})||p_{\theta}(\mathbf{z}_t^n|\mathbf{z}_{t-1}^n, \mathbf{d}^n)))
\label{eq:fullobj}
\end{split}
\end{equation}

\subsection{Functional Form}
As depicted in Figures \ref{fig:genmodel} and \ref{fig:infmodel}, there are a number of functions with learnable parameters in the generative and inference models of VDSM.

\textbf{Encoder and Static Factors}:  The encoder functions $\mbox{ENC}_{\{s, z\}}(.)$ (pink block in Figure \ref{fig:infmodel}) are used to infer the static and time varying embeddings of the images at each timestep. It comprises 5 layers of anti-aliased convolution \cite{Zhang2019conv} downsampling, as well as two separate fully-connected embedding layers with non-linear activations. The identity embedding is averaged over a sequence and fed through a fully-connected layer $f_s(.)$ (light gray block in Figure \ref{fig:infmodel}) which is used to infer the identity/static factor for a sequence $\mathbf{s}^n$.

\textbf{Dynamics Layer and LSTMs: } The Dynamics layer (red block Figure \ref{fig:infmodel}) comprises $f_{d\mu}(.)$ and $f_{d\sigma}(.)$ which are fully-connected neural network layers with non-linear activations used to infer the location and scale of the dynamics/action factor $\mathbf{d}^n$ and are fed with the last hidden state from the encoder bi-LSTM. It therefore represents a bottleneck inside a seq2seq network. The seq2seq network's uni-directional LSTM decoder uses the inferred $\mathbf{d}^n$ as the initial hidden and initial cell states, and produces per-timestep hidden states $\bar{\mathbf{h}}_t^n$ which are fed to the combiner.

\textbf{The Combiner}: The Combiner (yellow block in Figure~\ref{fig:infmodel}) is used to infer the current latent pose factor at time $t$ given the current hidden state from the decoder LSTM, the previous latent pose factor, and the dynamics: $\mathbf{z}_t^n|\mathbf{z}_{t-1}^n, \bar{\mathbf{h}^n_t}$. The function is parameterized as follows:
 
 % there was a typo in the original submission which incorrectly define the input to the combiner - there is no d, and therefore no concat operation
 
 \begin{equation}
\begin{split}
\hat{\mathbf{h}}_t^n = \{\bar{\mathbf{h}}_t^n, \mathbf{d}^n\}\\
\mathbf{c}_t^n = 0.5(\mbox{tanh}(f_7(\mathbf{z}^n_{t-1})) + \hat{\mathbf{h}}_t^n)\\
\mbox{loc-comb}_t^n = f_8(\mathbf{c}_t^n)\\
\mbox{scale-comb}_t^n = \log(1 + \mbox{exp}(f_{9}(\mathbf{c}_t^n)))\\
\mathbf{z}_t^n \sim \mathcal{N}(\mbox{loc-comb}_t^n, \mbox{scale-comb}_t^n)
\end{split}
\end{equation}

Where $\{\bar{\mathbf{h}}_t^n, \mathbf{d}^n\}$ is the concatenation of the LSTM decoder hidden state and the dynamics, and functions $f_{7-9}(.)$ are fully-connected neural network layers with non-linear activations. During development we explored the use of Inverse Autogressive Flows (IAFs) to increase the expressivity of the approximating posterior. IAFs are an adapted form of Normalizing Flow \cite{Kobyzev2020, rezende} designed for efficient transformation of simple posterior approximating distributions. IAFs leverage an invertible transformation function with a particularly tractable Jacobian determinant \cite{rezende, Kobyzev2020, Kingma2016}. However, at least in our particular experiments, we found that the concomitant increase in computational complexity was not justified by the negligible performance increase.

\textbf{Transition Network: } Similar to the state-space transition model in \cite{Krishnan2015}, the transition network (green block Figure \ref{fig:genmodel}) parameterizes the latent time-varying factors $\mathbf{z}^n_t$ with functions $\mu_{tr}(.), \sigma_{tr}(.)$, each with the following parameterization:

\vspace{-2mm}
\begin{equation}
\begin{split}
\mathbf{g}_t^n = \mbox{sigmoid}(f_2(\mbox{ReLU}(f_1(\{\mathbf{z}_{t-1}^n, \mathbf{d}^n\}))))\\
\mathbf{h}_t^n = f_4(\mbox{ReLU}(f_3(\{\mathbf{z}_{t-1}^n, \mathbf{d}^n\})))\\
\mbox{loc-trans}_t^n = \mathbf{g}_t\circ \mathbf{h}_t^n + (1-\mathbf{g}_t^n)\circ f_5(\{\mathbf{z}_{t-1}^n, \mathbf{d}^n\}) \\
\mbox{scale-trans}_t^n = \log(1 + \mbox{exp}(f_6(\mbox{ReLU}(\mathbf{h}_t^n))))
\end{split}
\end{equation}

where $\circ$ indicates the Hadamard/elementwise product, $\{\mathbf{z}_{t-1}^n, \mathbf{d}^n\}$ is the concatenation of the previous pose vector with the dynamics vector for sequence $n$, and the functions $f_{1-6}(.)$ are fully-connected neural network layers with non-linear activations. This network is used as part of the generation process to generate successive latent pose factors $\mathbf{z}_t^n|\mathbf{z}^n_{t-1}$. 

   \begin{figure*}
\centering
\includegraphics[width=0.8\linewidth]{ 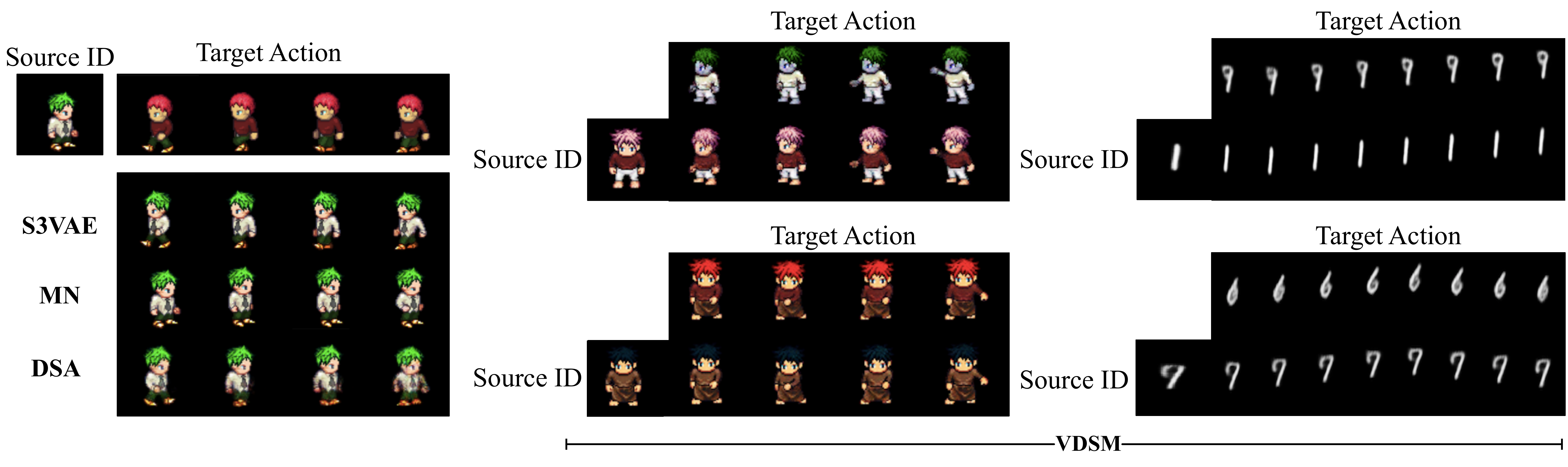}
\caption{\textbf{Left}: Sprites dataset \cite{Li2018b} action transfer examples for S3VAE \cite{Zhu2020}, MN (MonkeyNet) \cite{Siarohin2019MonkeyNet} and DSA \cite{Li2018b} (results adapted from \cite{Zhu2020}). \textbf{Middle} and \textbf{right}: VDSM action transfer on Sprites and moving MNIST datasets \cite{Srivastava2016} respectively. `Target Action' indicates the desired action, whilst `Source ID' indicates the identity to which the action is transferred during sequence generation.}
\label{fig:mixed_results}
\end{figure*}

\textbf{Mixture of Experts Decoder: } The Mixture of Experts (MoE) \cite{Zhang2018, Shazeer2017, Jordan1993, Eigen2013} decoder $\mbox{DEC}_{MoE}(.)$ (blue block in Figure \ref{fig:genmodel}) constitutes a bank of $\mathrm{N}_s$ upsampling convolution image decoders. Each decoder comprise a series of 2D transpose convolution and Leaky ReLU activation layers. The weights and biases for the transpose convolution operations of the bank of MoE decoders are \textit{blended} according to a weighted average, where the weighting is determined by the identity mixture parameter $\hat{\mathbf{s}}^n$. In addition, $\mathbf{s}^n$ (which is $\hat{\mathbf{s}}^n$ before temperature annealing and the softmax operation) is concatenated with $\mathbf{z}_t^n$ and fed to the input of the decoder. As the temperature parameter $\tau_s$ is annealed during training, the mixture parameter becomes more `peaked' resulting in a tendency to select individual decoders from the bank, rather than mix between them. Each decoder in the bank is therefore given the opportunity to specialize in reconstructing a particular identity, whilst also leveraging information encoded in the weights in the entire bank. The MoE decoder is used as part of the generation process to generate the images $\mathbf{x}_t^n | \mathbf{z}_t^n, \mathbf{s}^n $.

% \begin{equation}
%     \log q_\phi(\mathbf{z}^J | \mathbf{x}) = \log q_{\phi}(\mathbf{z}^{j=0}|\mathbf{x}) - \sum_j^J \log \mbox{det} \abs{\frac{d\mathbf{z}^j}{d\mathbf{z}^{j-1}}}
% \end{equation}
% where we have temporarily overloaded the notation by using superscript $j$ to indicate the transformation number in the NF. IAFs leverage an invertible transformation function with a particularly tractable Jacobian determinant (see references for further details) \cite{rezende, Kobyzev2020, Kingma2016}.

\subsection{Training and Testing VDSM}
 \textbf{Training:} Training is split into two stages. Without two stages, the network tends to push both pose and identity information into $\mathbf{d}$ to avoid the averaging operation associated with $\mathbf{s}$. We start by \textbf{pretraining} the encoder $\mbox{ENC}_{\{s,z\}}$, MoE decoder $\mbox{DEC}_{MoE}$ and the Static Factors Layer $f_s$ without modeling the autoregressive structure of the data (this is illustrated in the gray shaded regions of Figures \ref{fig:genmodel} and \ref{fig:infmodel}. In other words, we begin by treating the data as non-sequential, and group images from the same individuals/identities into batches in order to facilitate the inference of the identity factor $\mathbf{s}^n$ and i.i.d. $\mathbf{z}_t^n$. During this first stage, the objective function is reduced to the following:
\vspace{-2.5mm}
 \begin{equation}
 \begin{split}
\mathcal{L}(\mathbf{x}^n_{1:T_n}; (\theta', \phi')) = 
\frac{1}{N} \sum_{i=1}^{T}(\mathbb{E}_{q_{\phi'}(\mathbf{z}_i, \mathbf{s}| \mathbf{x}_i)}[\log p_{\theta'}(\mathbf{x}_i | \mathbf{z}_i, \mathbf{s})]\\ 
- \lambda_z \mbox{KL}[q_{\phi'}(\mathbf{z}_i | \mathbf{x}_i) \| p_{\theta'}(\mathbf{z})]) 
- \lambda_s \mbox{KL}[q_{\phi'}(\mathbf{s} | \mathbf{x}_i) \| p_{\theta'}(\mathbf{s})])
 \end{split}
\label{eq:ELBOpretrain}
\end{equation}

 Here we deliberately use subscript $i$ rather than $t$ to emphasize that the images need not be in any sequential order, they just need to be grouped according to the same identity. Furthermore, we indicate that we are only training a subset of the inference and generation parameters with $(\theta', \phi')$. During pretraining, the KL weight on the static factors is inversely annealed $\lambda_s = \{ 0.1, ..., 1.0 \}$ (i.e. low to high) with respect to the KL weight on the pose factor $\lambda_z = \{ 30, ..., 1.0 \}$ (i.e. high to low), with the number of steps equal to the number of training epochs. This forces information to flow through the static factor $\mathbf{s}^n$ because the high weight $\lambda_z$ pushes the time varying factors $\mathbf{z}_i$ to zero. The static factor is computed as an average over embeddings and as such, it is forced to represent the information which is consistent over the group of images (i.e. the identity). Simultaneously, the temperature parameter $\tau_{s}$ is gradually increased, resulting in a shift from uniform MoE decoder blending weights, to increasing specialization. Then, as the weight $\lambda_z$ falls, the pose factors $\mathbf{z}_i$ start to become useful in encoding information that varies across the groups of images of the same individual. As such, the pretraining enables us to learn the two most highly-parameterized deterministic functions used for mapping images to latent factors and vice versa: the encoder and the MoE decoder. It is worth noting that the pretraining stage yields a valuable model in its own right, that disentangles identity from pose. For instance, in Figure~\ref{fig:pretrained_MUG_swap}, the expression can be swapped with any ID and vice versa.

The \textbf{second stage of training} is concerned with disentangling the sequence dynamics, and introduces the seq2seq encoder and decoder LSTMs, the Combiner, and the transition network. The training objective is given by the full ELBO in Eq. \ref{eq:fullobj}. The weights and biases for the encoder and MoE decoder are all frozen at the end of pretraining, apart from the final layers preceding the parameterizations of $\mathbf{s}^n$ and $\mathbf{z}_t^n$ which are allowed to vary during this second stage. Allowing these layers to be fine-tuned is important in ensuring that the model learns a valid ELBO during the sequential modeling stage (i.e. by allowing the parameters to vary we facilitate variational inference). Annealing is used with $\lambda_z = \{0.1, ..., 1.0\}$ to improve inference of $\mathbf{z}_t$ and to help prevent posterior collapse during training \cite{disentanglement}. 

 To infer $\mathbf{s}^n$, which is primarily learned during the first stage of training (but fine-tuned in the second stage), we only need to be able to group images of the same individual (and these may or may not be from the same sequence). For $\mathbf{d}^n$ the images need to be from sequences, but no supervision is required. This allows us to sample multiple images of the same individual, without needing labels for which individual is being sampled. So long as there is, on average, some variation across identity between groups of images (or between sequences), the network can infer the identity of the individual. VDSM is trained using the Adam \cite{adam} optimizer with the Stochastic Variational Inference algorithm in the Pyro probabilistic programming language \cite{Bingham2019pyro}. Minimal hyperparameter tuning was undertaken.\footnote{Network details can be found in supplementary material.}

\textbf{Testing:} At test time, the generative model may either be used unconditionally (where the factors $\mathbf{s}^n, \mathbf{d}^n$, and $\mathbf{z}^n_t$ are sampled from their priors), or be used conditioned on some initial state. In the latter case, the inference network may be used to derive $\mathbf{s}^n, \mathbf{d}^n, \mathbf{h}^n_{t=1}$, and $\mathbf{z}^n_{t=1}$ from a sample sequence, or even a single image. The generative model is then conditioned on these initial factors, and further sampling is undertaken according to Figure \ref{fig:genmodel}. The benefit of this flexible structure is that it allows different factors to be swapped or controlled independently in order to mix any identity with any action dynamics.

\begin{figure}
\centering
\includegraphics[width=0.85\linewidth]{ 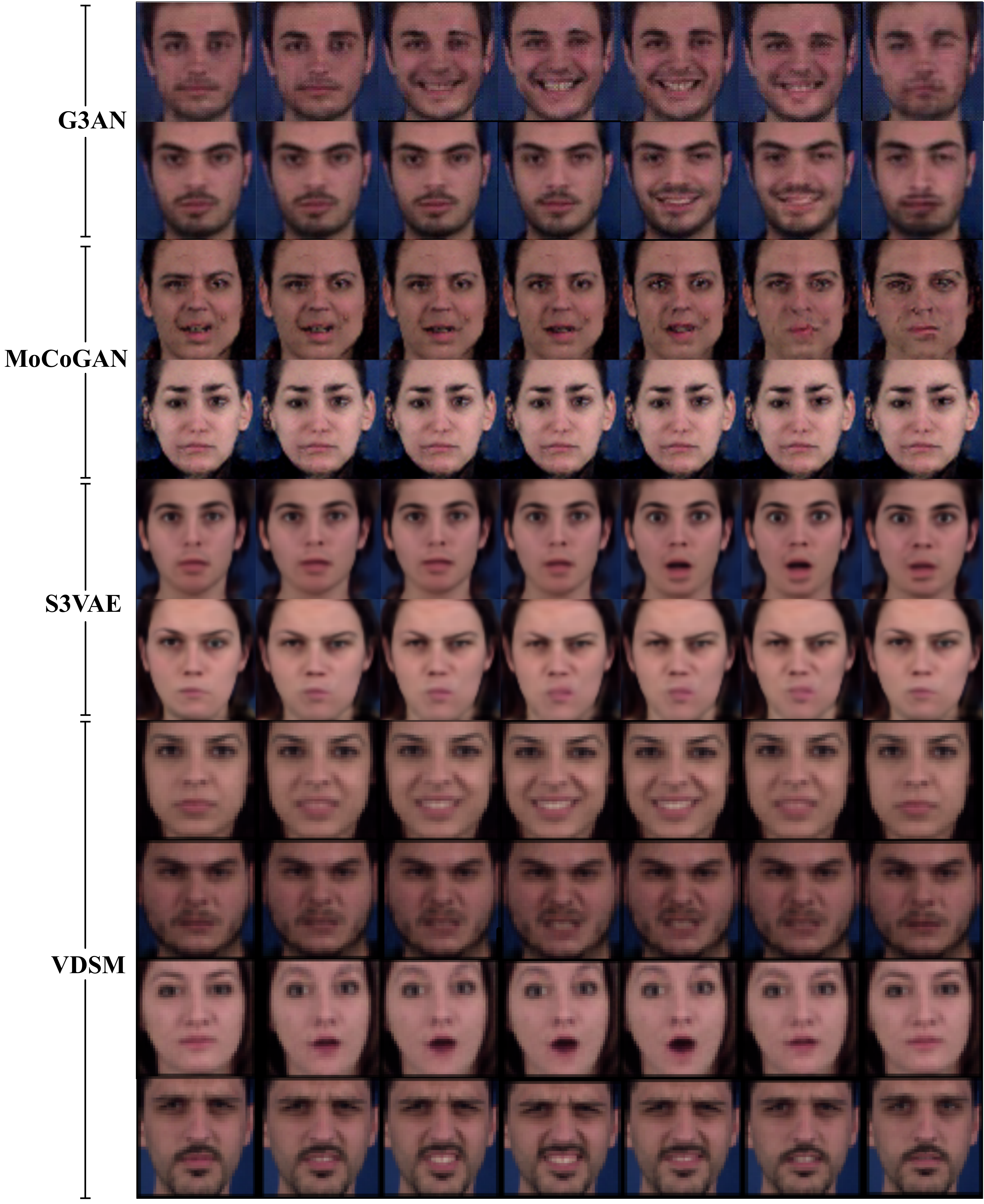}
\caption{Downsampled sequence generation samples from G3AN \cite{Wang2020b}, MoCoGAN \cite{Tulyakov2017}, S3VAE \cite{Zhu2020}, and VDSM (ours). Figure adapted from \cite{Zhu2020, Wang2020b}.}
\label{fig:MUG_seq_gen}
\end{figure}

\section{Experiments}
 VDSM is evaluated on four datasets: MUG \cite{Aifanti2010MUG}; colored Sprites \cite{Li2018b}; synthetic swinging pendula; moving MNIST (similar to \cite{Srivastava2016}). MUG \cite{Aifanti2010MUG} comprises 3528 videos of 52 individuals performing different facial expressions for anger, disgust, fear, happiness, sadness, and surprise. The performances vary in length, and were downsampled by a factor of two (to approximately 8 fps) to improve data efficiency. The images were aligned using OpenFace \cite{openface2}, centre cropped, and resized to 64x64 (similar to \cite{Tulyakov2017, Zhu2020}). Random segments of length 16 frames were sampled from the sequences for training, and a 15\% holdout set was used for testing.

 The Sprites dataset comprises 64x64 sequences of cartoon characters from the Universal LPC SpriteSheet Character Generator performing 3, 8-frame long action sequences (spellcast, slashing, and walking), from 3 viewing angles. Following \cite{Li2018b} we create characters with 7 body types, 4 shirts, 8 hairstyles, and 5 pants, resulting in 1120 identities. A 10\% holdout set was used for testing. The moving MNIST dataset \cite{Srivastava2016} comprises 55,000 16-frame long sequences of randomly sampled MNIST digits \cite{MNIST} moving in random trajectories. A 10\% holdout set was used for testing. Finally, the pendulum dataset represents a synthetic dataset which comprises seven different colored pendula swinging at two speeds (fast and slow).
 
  We evaluate quantitatively in terms of (1) Fr\'echet Inception Distance (FID) \cite{Heusel2017FID, Borji2018GAN} using a 3D ResNeXt 101 \cite{Xie2017ResNeXt, Hara2017} pre-trained on Kinetics \cite{Kay2017Kinetics} as per \cite{Wang2020b}, (2) consistency at identity and action classification between real and generated sequences \cite{Li2018, Zhu2020}, and (3) in terms of identity and action classification score using the separate identity and dynamics embeddings.\footnote{Additional results may be found in supplementary material.} For (2), the consistency is measured by comparing the predictions of a classifier trained to predict ground truth factors from real images with predictions from sampled images. If the quality of the images is high and the network is encoding information about the identity and action, then the predictions for real and generated images should be close. For (3) we expect that disentangled identity and action embeddings to be informative for predicting their respective factors, but not informative for predicting each other's factors. 
  %In other words, we hope that the identity embedding is useful for predicting identity, but not for predicting action.
  Finally, we evaluate quantitatively, in terms of identity swapping, dynamics swapping, and sequence generation quality.

 \begin{figure}
\centering
\includegraphics[width=.9\linewidth]{ 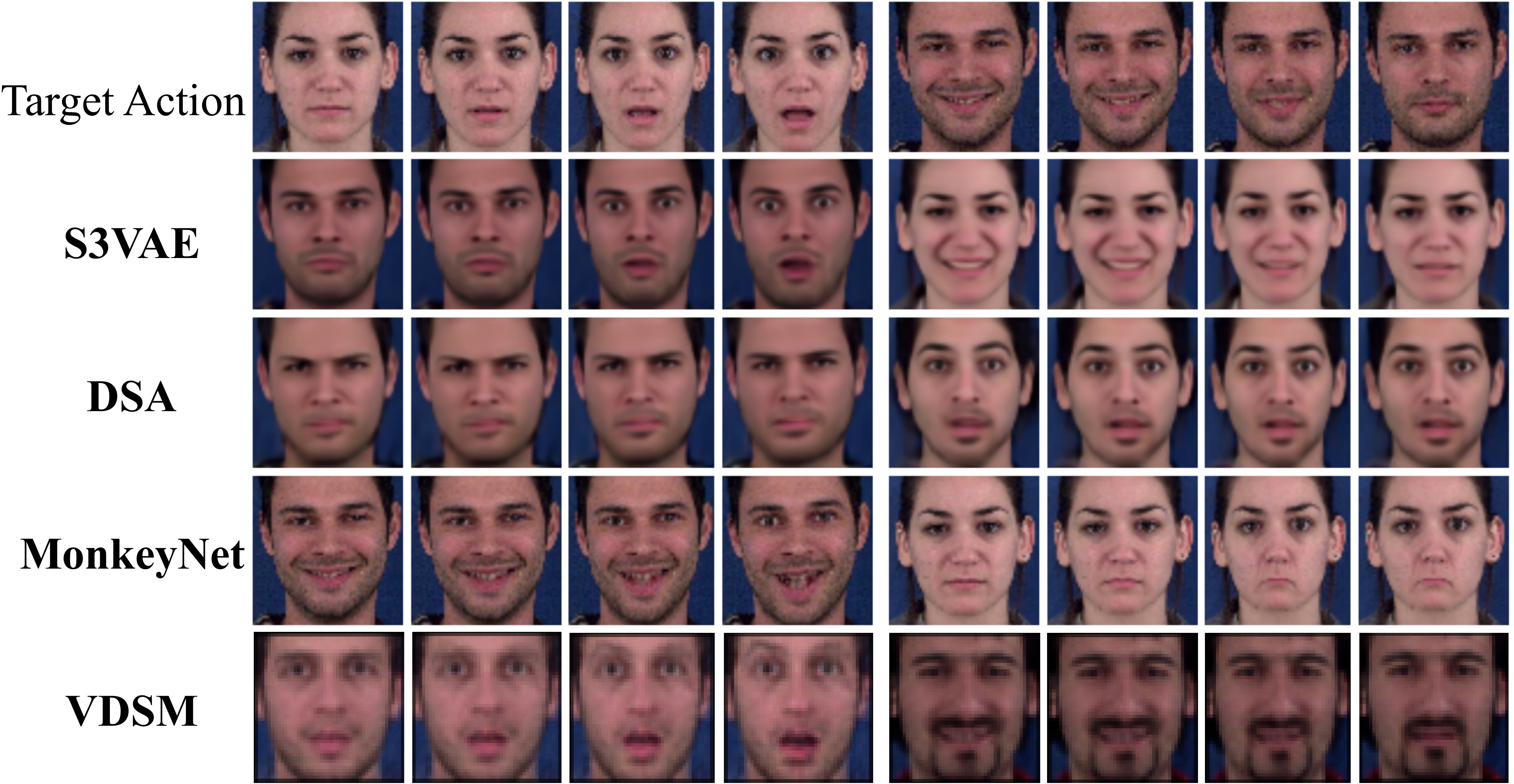}
\caption{Action transfer on the MUG dataset \cite{Aifanti2010MUG} for S3VAE \cite{Zhu2020}, DSA \cite{Li2018b}, MonkeyNet \cite{Siarohin2019MonkeyNet}, and VDSM (ours). Figure adapted from \cite{Zhu2020}.}
\label{fig:MUG_action_transfer}
\end{figure}

\begin{table}[]
\small
\begin{tabular}{llll} \hline
\textbf{Attr.} & \textbf{Acc. Cons}. $\uparrow$ & \textbf{ID} \textbf{Acc}. & \textbf{Dyn. Acc}.  \\ \hline
Body (Sprites) & 99.67 & 98.67 $\uparrow$& 52.67 $\downarrow$\\
Shirt (Sprites)& 99.67 & 99.67 $\uparrow$& 47.33 $\downarrow$ \\
Pant (Sprites)& 99.00 & 97.67 $\uparrow$& 31.67 $\downarrow$\\
Hair (Sprites)& 100.00 & 99.33 $\uparrow$& 21.00 $\downarrow$\\
Action (Sprites) & 99.33 & 45.00 $\downarrow$ & 99.33 $\uparrow$\\
\hline
ID (MUG) & 98.67 & 99.00 $\uparrow$ & 65.67 $\downarrow$\\ 
Action (MUG) & 88.00 & 57.33 $\downarrow$ &  83.33 $\uparrow$\\ 
\hline \hline
\end{tabular}
\caption{For each ground-truth attribute of Sprites and MUG, this table shows a breakdown of VDSM's consistency (Acc. Cons.) and disentanglement accuracy scores in percent using the separate identity (ID) and dynamics/action (Dyn.) embeddings. Arrows indicate whether a higher $\uparrow$ or lower $\downarrow$ score is preferred.}  
\label{tab:disent}
\end{table}

\begin{table}[]
\small
\begin{tabular}{lllll} \hline
\textbf{Method} & Acc. \textbf{Sprites} & Acc. \textbf{MUG} & $H(y)$ &  $H(y|x)$ \\ \hline
VDSM  & \textbf{99.53} & \textbf{93.33} & \textbf{2.21} & 0.203   \\
S3VAE \cite{Zhu2020}& \textbf{99.49} & 70.51& 1.760& \textbf{0.135}\\
DSA \cite{Li2018b} & 90.73 & 54.29 & 1.657 & 0.374\\
MoCo \cite{Tulyakov2017} & 92.89 & 63.12 &1.721 & 0.183\\
\hline \hline
\end{tabular}
\caption{Results for accuracy consistency (averaged across attributes) between different methods for the Sprites and MUG datasets, and results for Inter-Entropy $H(y)$ (higher is better) and Intra-Entropy $H(y|x)$ (lower is better) \cite{He2018} on the MUG dataset.}
\label{tab:consall}
\end{table}

  \subsection{Quantitative Evaluation}
 
  Quantitative results were obtained for 1000 samples from the test sets. The results in Table \ref{tab:disent} show that the ID embedding is predictive of ID or ID-related attributes (acc. $>97\%$ for Sprites), but not of action (acc.~$=45\%$ for Sprites). Similarly, the dynamics embedding was highly predictive of action (acc. $>99\%$ for Sprites) but not of ID-related attributes (acc. $<53\%$ for Sprites). Interestingly, the ID could be classified using the dynamics embedding for MUG (acc. 66\%). We believe this may be because individuals exhibit unique action dynamics that make them identifiable. This behavior may have relevant application elsewhere (\textit{e.g.} gait recognition). Future work should explore whether this behavior is exhibited by other methods. 
  
  The breakdown of VDSM's accuracy consistency scores in Table \ref{tab:disent} demonstrates that there is high corroboration between action and ID-related attribute classifier predictions for real and generated images. Indeed, the results for VDSM exceed those from competing methods. In particular, see the results in Table \ref{tab:consall} where MUG performance was~$>30\%$ higher than the best competing method (S3VAE) despite this method using additional supervision (\textit{e.g.} optical flow). Table \ref{tab:consall} also shows the results for Inter- and Intra-Entropy \cite{He2018} for the same classifier used for the disentanglement evaluation. These are computed as $H(y) = -\sum_y p(y) \log p(y)$ and $H(y|x) = -\sum_y p(y|x) \log p(y|x)$ respectively, where $y$ is the predicted attribute label, and $x$ is the given sequence. The inter- and intra-entropies are averaged across identity and action prediction results. The results for inter-entropy provide a proxy for diversity, and indicate that, in generation mode, VDSM produces the most diverse samples. The results for intra-entropy demonstrate competitive performance with another unsupervised method, DSA. However, together with the disentanglement and accuracy results in Table \ref{tab:consall} suggest that VDSM's primary strength is superior, unsupervised, disentangled representation learning. 
  
   \begin{figure}[!h]
\centering
\includegraphics[width=0.6\linewidth]{ 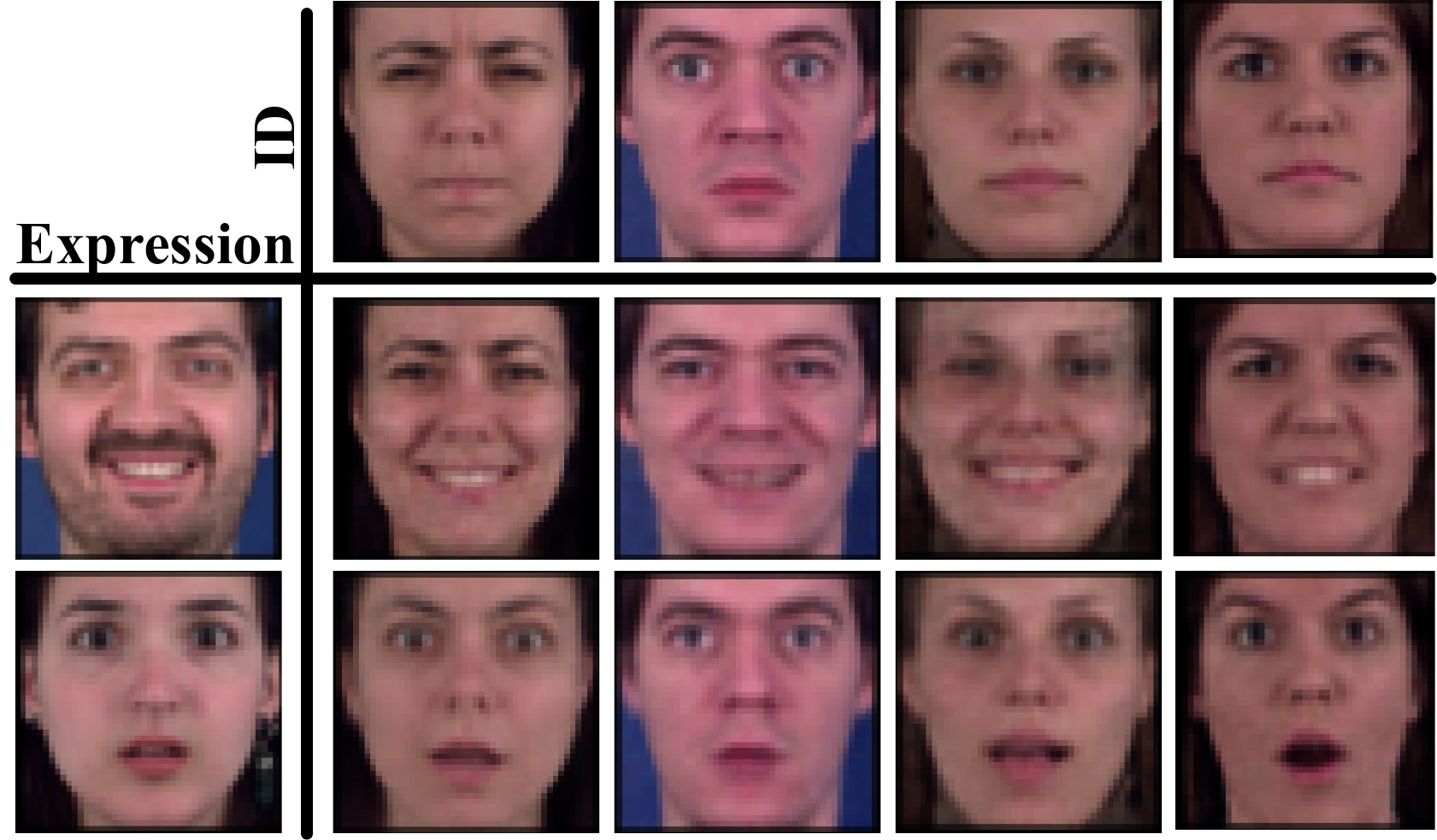}
\caption{Identity/pose transfer using the pre-trained (non-sequential) components of VDSM on the MUG dataset \cite{Aifanti2010MUG}.  }
\label{fig:pretrained_MUG_swap}
\end{figure}

  FID scores for MUG are shown in Table \ref{tab:FID}. For this evaluation 1000 samples were generated firstly by using $\mathbf{d}^n$ and $\mathbf{s}^n$ embeddings derived from real sequences in the test set (denoted `reconstruction' in the table), and secondly by generating images unconditionally from the priors (denoted `generation' in the table). In both case, the FID substantially improves upon competing adversarial methods.
 
 \begin{table}[]
 \centering
\small
\begin{tabular}{lll} \hline
  \textbf{Method}      &\textbf{FID}    \\ \hline
VDSM (ours, generation)   & \textbf{44.12}       \\
VDSM (ours, reconstruction)   & \textbf{17.64}       \\
G3AN \cite{Wang2020b}       & 67.12         \\
MoCo \cite{Tulyakov2017}      & 87.11     \\
VGAN  \cite{Vondrick2016VGAN}       & 160.76   \\
TGAN \cite{Saito2017TGAN}       & 97.07    \\ \hline \hline
\end{tabular}
\caption{FIDs for MUG. We include VDSM in generation and reconstruction modes, where samples are drawn unconditionally from priors or are drawn from approximate posteriors derived from test set sequences, respectively.}
\label{tab:FID}
\end{table}

 \subsection{Qualitative Evaluation}
 Figures \ref{fig:sprites_action_swap} and \ref{fig:mixed_results} illustrate VDSM's action transfer performance on the Sprites dataset, and some comparisons are included in the leftmost part of Figure \ref{fig:mixed_results}. It can be seen that VDSM facilitates disentangled transfer, with clean separation of time varying and time static information. In contrast, DSA and MonkeyNet struggle to transfer the action, and S3VAE requires auxiliary information to achieve comparably (\textit{e.g.} optical flow). The rightmost part of Fig. \ref{fig:mixed_results} shows action transfer of VDSM on the moving MNIST dataset.
 
 Figure \ref{fig:MUG_seq_gen} compares VDSM's sequence generation performance against G3AN, MocoGAN, and S3VAE, where it can be seen that VDSM outperforms the adversarial methods, and performs comparably to S3VAE, despite not requiring additional information (\textit{e.g.} optical flow).\footnote{Sequences are downsampled - for full sequences see supplementary material.} Figure \ref{fig:MUG_action_transfer} compares the action transfer performance of VDSM against DSA, S3VAE and MonkeyNet, again highlighting that VDSM is comparable to S3VAE despite not requiring additional information during training (\textit{e.g.} optical flow). Figure \ref{fig:pretrained_MUG_swap} demonstrates the \textit{pre-trained} components of VDSM performing identity and expression transfer, illustrating how the model can disentangle pose and identity without needing to model sequences. 
 
  Results on the synthetic pendulum dataset are shown in Figure 11 in the supplementary material. These results demonstrate that, even though VDSM was only trained on sequence segments which were 16 frames long, the network can generate sequences following the dynamics of a pendulum with an arbitrary number of frames (40 frames are shown in the figure). The figure also demonstrates action swapping, where the dynamics of one pendulum is transferred to another.

\section{Summary}
We presented VDSM, an unsupervised state-space model for video disentanglement. VDSM incorporates a range of inductive biases including a hierarchical latent structure, a dynamic prior, a seq2seq network, and a mixture of experts decoder. The evaluation demonstrated that the network informatively embeds and disentangles static and time-varying factors, as well as generating quality video many frames into the future. VDSM's performance matches or exceeds that of methods requiring additional supervision (such as optical flow) or adversarial training.

{\small
\bibliographystyle{ieee_fullname}
\bibliography{cvpr}
}

\clearpage

\appendix

\section{Supplementary Material}

\begin{figure}[h!]
\centering
\includegraphics[width=1\linewidth]{ 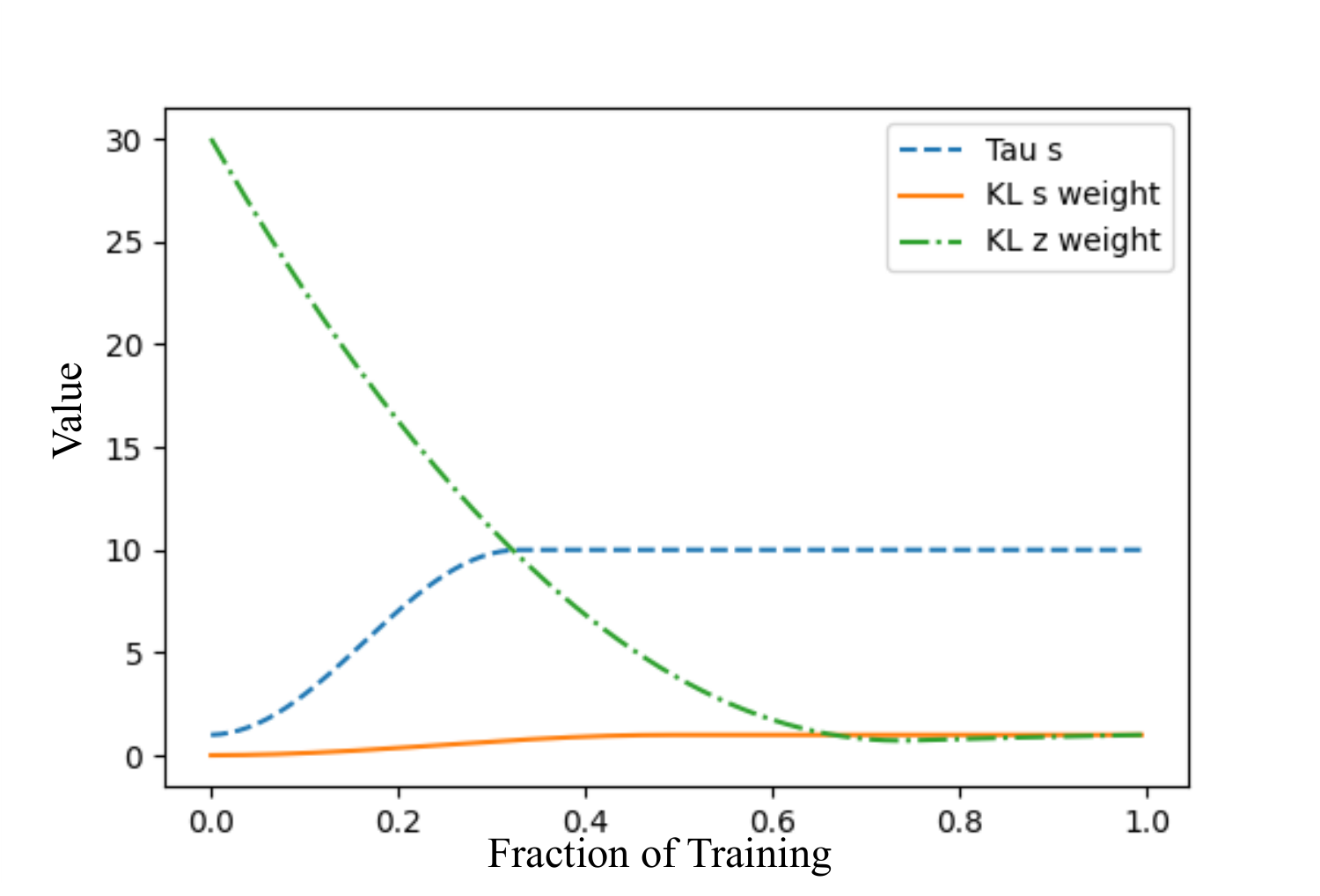}
\caption{Annealing schedules and temperature schedules for VDSM during pretraining. Shows the weight $\lambda_z$ on the KL divergence for the time varying pose factor $\mathbf{z}_t^n$, the weight $\lambda_s$ on the KL divergence for the identity factor $\mathbf{s}^n$, and the temperature $\tau_s$ for the identity factor. }
\label{fig:annealingpre}
\end{figure}

\begin{figure}[h!]
\centering
\includegraphics[width=1\linewidth]{ 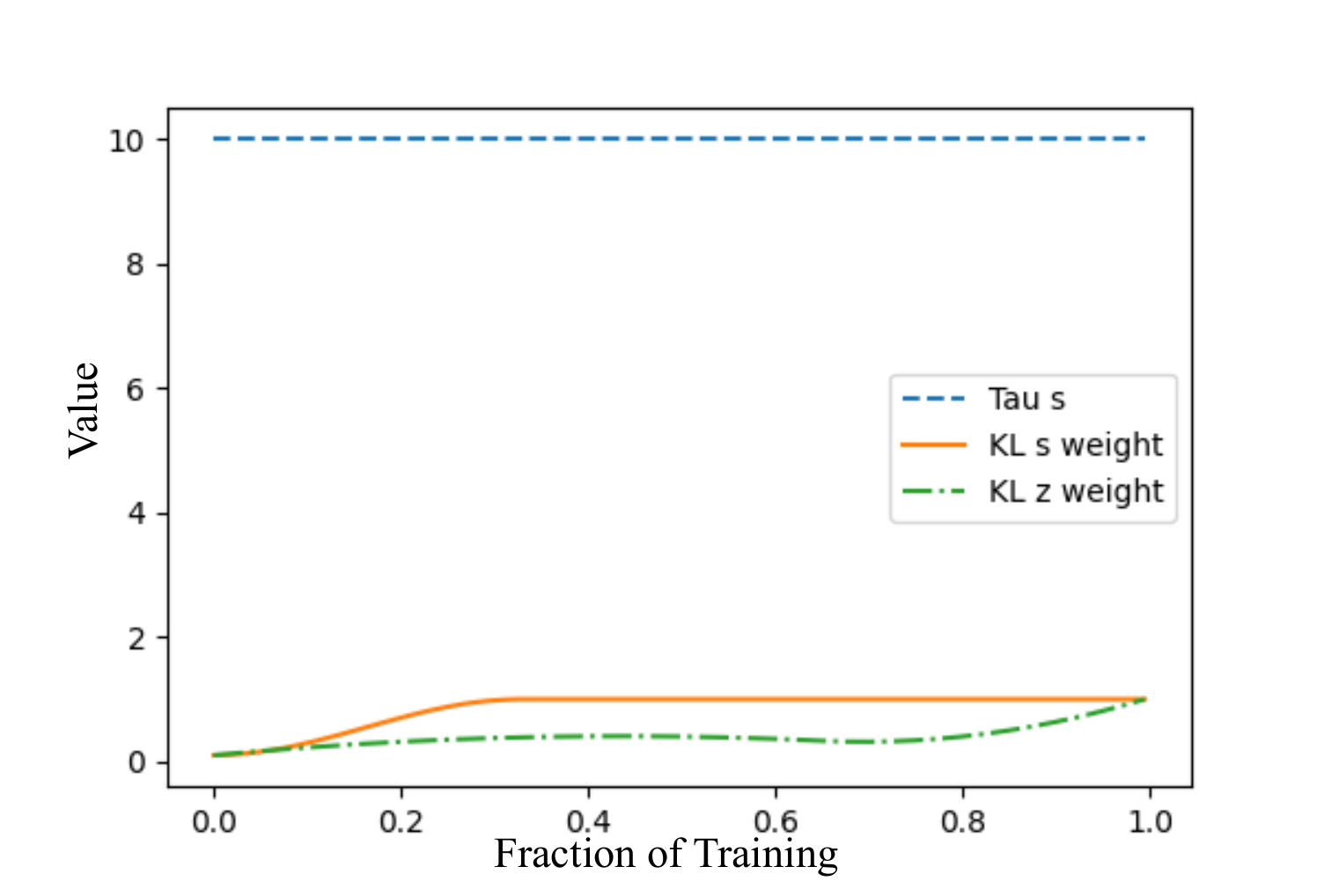}
\caption{Annealing schedules and temperature schedules for VDSM during training of the sequential components. Shows the weight $\lambda_z$ on the KL divergence for the time varying pose factor $\mathbf{z}_t^n$, the weight $\lambda_s$ on the KL divergence for the identity factor $\mathbf{s}^n$ (constant during this phase of training), and the temperature $\tau_s$ for the identity factor. }
\label{fig:annealingseq}
\end{figure}

\begin{table*}[h!]
\begin{tabular}{lccccccccc}\hline
Dataset & lr-Pre & lr-Seq & Epochs-Pre & Epochs-Seq & BS-Pre & BS-seq & Seq Len & BPE-Pre & BPE-Seq \\ \hline
MUG     & 1e-3   & 1e-3   & 250        & 200        & 20     & 20     & 20      & 50      & 50      \\
Sprites & 8e-3   & 1e-3   & 300        & 200        & 20     & 20     & 8       & 50      & 50      \\
MMNIST  & 1e-3   & 1e-3   & 300        & 200        & 20     & 30     & 16      & 50      & 50      \\
Pendula & 5e-4   & 1e-3   & 200        & 100        & 50     & 20     & 16      & 50      & 50     \\ \hline \hline
\end{tabular}
\caption{VDSM hyperparameter settings for each dataset. `Pre' and `Seq' refer to the pre-training and sequence training respectively. `BS' is batch size, `lr' is learning rate, and `BPE' is the number of batches per epoch.}
\label{tab:hypers}
\end{table*}

\section{Code and Qualitative Video Samples}
Code and example video can be found via the following URL: \url{https://github.com/matthewvowels1/DisentanglingSequences} as well as in the `samples' folder in the supplementary material.\footnote{These are .gif files which may need to be viewed in (e.g.) an internet browser for animation.}

\section{Overview of Supplementary Material}
This supplementary material provides additional information and results for the work titled `VDSM: Unsupervised Video Disentanglement with State-Space Modeling and Deep Mixtures of Experts'. We first provide details about the network architecture, training details, run-time estimates, and briefly discuss the results of a simple ablation experiment. Further results are given in the supplementary for the official CVF CVPR paper. We then provide a derivation for the ELBO presented in Equation 4 in the main text, and finally present a range of qualitative results for the MUG \cite{Aifanti2010MUG}, Sprites \cite{Li2018b}, moving MNIST \cite{Srivastava2016} and synthetic pendulum dataset.

\section{Network Architecture}
The network was implementation using a combination of Pytorch \cite{pytorch} and Pyro \cite{Bingham2019pyro}, and the code has been included as part of the supplementary material. Various relevant hyperparameters and dimensionalities are shown in Table \ref{tab:hypers} and \ref{tab:dims}.

\textbf{Encoder and Static Factors}:  
The encoder comprises the following blocks:

\noindent [Conv2D(32,4,1), LeakyReLU, BlurPool], [Conv2D(32,4,2), LeakyReLU, BlurPool], [Conv2D(32,4,2), LeakyReLU, BlurPool], [Conv2D(64,4,2), LeakyReLU, BlurPool], [Conv2D(64,4,2), \hspace{12mm} LeakyReLU] 

\noindent where Conv2D($x, y, z$) is the convolution operation with $x, y, z$ being the number of output filters, the kernel size, and the stride, respectively. The first block (only) has padding of 1. LeakyReLU is the leaky rectified linear unit \cite{leakyrelu}, and blur pool enables anti-aliased downsampling \cite{Zhang2019conv}. The output is reshaped and fed to separate two consequetive fully-connected layers [FC(256, 128), FC(128, 2$\kappa_s$)] (where the two arguments are the number of input and output neurons) to yield the embeddings for $\mathbf{s}^n$ (the identity), and fed to a single fully connected layer [FC(256, 2$\kappa_z$] to yield the embeddings for $\mathbf{z}_t^n$ (the time varying components). These embeddings are split into two to yield the location and scale parameters of the Normal distributions used to model the two factors.

\begin{table}[h!]
\begin{tabular}{llllcc} \hline
\textit{Dataset}   & $\kappa_z$ & $\kappa_s$ & $\kappa_d$ & \textit{RNN Layers} & \textit{RNN dim.} \\ \hline 
MUG       & 30      & 15      & 50      & 3          & 512         \\
Sprites   & 30      & 40      & 50      & 3          & 512         \\
MMNIST & 30      & 12      & 50      & 3          & 512         \\
Pendula   & 30      & 8       & 50      & 3          & 512        \\ \hline\hline
\end{tabular}
\caption{$\kappa_z$, $\kappa_s$, and $\kappa_d$ are the dimensionalities of the pose, identity, and dynamics/action latent factors, respectively. Both the bi-LSTM encoder and the uni-directional LSTM decoder have the same number of layers and hidden dimensions (3 and 512, respectively).}
\label{tab:dims}
\end{table}

\textbf{Dynamics Layer, LSTMs and Combiner:} The seq2seq encoder is a bi-LSTM, and the decoder is a uni-directional LSTM, each with settings listed in Table \ref{tab:dims}. The output of the bi-LSTM is a hidden representation with a dimensionality equal to $\mbox{RNN}_{dim} \times \mbox{RNN}_{layers} \times 2$. This hidden representation is fed into the full-connected dynamics layer with output dimensionality $2\times \kappa_d$, and is split in half to yield the location and scale of $\mathbf{d}^n$. The RNN decoder outputs per-timepoint vectors which are fed through a fully-connected layer $\mbox{FC}(\mbox{RNN}_{dim} \times 2, 2 \times \kappa_z)$. The intermediary hidden size of the combiner network is 512, and otherwise the parameter shapes of the fully connected layers in the combiner are determined by the dimensionalities of the inputs and the outputs of the function (i.e., the dimensionalities of $\mathbf{z}_t^n, \mathbf{h}_t^n$, and $\mathbf{d}^n$).

\textbf{Transition Network: } The transition network follows the structure described in the main paper. The intermediary hidden size used in the network is 64 and otherwise, like the combiner network, has fully connected layer weight sizes determined by the input and output dimensionalities of the function (i.e., the dimensionalities of $\mathbf{z}_t^n$ and $\mathbf{d}^n$).

\textbf{Mixture of Experts Decoder: } The Mixture of Experts (MoE) decoder (or generator) has $N_s=\kappa_s$ number of decoders which each follow this structure: 

[ConvTrans(1,0), LeakyReLU]

[ConvTrans(2,1), LeakyReLU]

[ConvTrans(2,1), LeakyReLU]

[ConvTrans(2,1), LeakyReLU]

[ConvTrans(2,1), LeakyReLU] 

where ConvTrans is 2-dimensional transpose convolution operation. The weights and biases for the ConvTrans operations are blended using the $\mathbf{s}^n$ sample which is duplicated $T_n$ times and concatenate with the pose vector for decoding.

% \begin{figure*}[!h]
% \centering
% \includegraphics[width=1\linewidth]{ pendulum_results_1.png}
% \caption{VDSM-generated frames for the synthetic pendulum dataset. Left figure illustrates how VDSM can be used to generate video far into the future (40 frame shown). Right figure illustrates how VDSM can transfer the action/dynamics onto a new identity. For this we simply use $\mathbf{d}^n$ from the purple or red pendulum sequences and apply it to generate sequences for the orange and blue pendula. `Target' indicates the desired action, whilst `ID' indicates the identity to which the action is transferred during sequence generation.}
% \label{fig:pendulum}
% \end{figure*}

\textbf{Annealing Schedules: }
Although very little tuning was required, the schedules for annealing the weights on the KL terms in the objective do need to be considered. Figures \ref{fig:annealingpre} and \ref{fig:annealingseq} show the annealing schedules for pre-training and sequence training. The function describing the profile of the pretraining curves is sinusoidal, whereas for the sequential training $\lambda_z$ curve is derived using quadratic interpolation.

\section{Derivation of the Lower Bound}
The derivation follows the same process as in \cite{Krishnan2016}. We first present the factorization of the generative and inference models in Equations \ref{eq:genmod} and \ref{eq:infmod}, respectively (it may be useful to reference the DAGs in the main paper). The compact representation of the ELBO objective is then shown in Equation \ref{eq:elboreduced}, and its final form is shown in Equation \ref{eq:full}. The first line in Equation \ref{eq:full} is derived straightforwardly according to the factorization of Equations \ref{eq:genmod} and \ref{eq:infmod}. However, the second line s further attention, and relates to the time-dependent nature of the pose factor $\mathbf{z}^n_t$ and its dependence on the dynamics $\mathbf{d}^n$. Omitting the weighting factor $\lambda_z$, The second line can be compactly reduced to Equation \ref{eq:reduced}. Note that the derivation and equations have been presented in single column format for legibility.

\section{Additional Qualitative Results}
Additional, randomly sampled qualitative results can be found in supplementary material (see CVF CVPR version) beginning with samples from the swinging pendulum dataset, then moving MNIST, Sprites, and finally MUG.

\section{Ablation - Using a Single Decoder}
Whilst the encoder used to derive a compact representation from the video frames was of comparable complexity to alternative/competing methods, the complexity of the mixture-of-experts decoder is arguably much greater. This is because it essentially comprises a bank of decoders, each which their own set of trainable weights. The network derives a mixing coefficient (which tends towards a discrete categorical latent variable) that blends or selects from the bank of decoders. Even though only one blended set of weights from the complete bank of weights is used for any one sequence, there is a significantly larger number of possible decoder configurations owing to the use of mixing. 

We ran an additional experiment to explore what happens if we use only the inferred mixing coefficient as a latent variable alone, and do \textit{not} use a bank of decoder weights (i.e. just a single decoder). We found that the reduced model resulted in a complete failure of the model to disentangle identity from pose (\textit{i.e.} both factors were highly entangled, and identify/pose swapping was not possible). It is difficult to ascertain to what extent this failure is due to the reduction in complexity associated with the use of single set of decoder weights, and to what extent it has something to do with a difference in resulting optimization dynamics which lead to different convergence properties. One possible way to establish this would involve a full hyperparameter search over the reduced model (the one without the mixture of decoders) to understand whether it is possible to achieve convergence. We leave this to future work.

\section{Hardware and Run Times}
The model was trained an tested on a GPU (e.g. NVIDIA 2080Ti) driven by a 3.6GHz Intel I9-9900K CPU running Ubuntu 18.04. Using the Sprites dataset by way of example, pre-training (1st stage) took 15 seconds for each of the 300 epochs, completing in 75 minutes. Sequence training (2nd stage) took 43 seconds for each of the 200 epochs, completing in approximately 2 hours 20 minutes. It is worth noting that pretraining and sequence training was found to converge significantly faster - as few as 100 epochs and 80 epochs respectively - corresponding to a total training time of approximately 80 minutes. However, a limited hyperparameter space was explored for this work, and we leave detailed efficiency studies to future work. At inference time, it was found that a batch of 20 sequences could be generated and saved to disk in approximately 0.2 seconds. 

Using the code provided here: \url{https://github.com/DLHacks/mocogan} we ran a $64\times64\times3$ version of the Weizmann dataset \cite{Gorelick2007Weizman} to get an approximate training time comparison against MoCoGAN \cite{Tulyakov2017}. The default frame dimensions are $96\times96\times3$, and so it was first necessary to modify the generators, discriminators, and dataset, accordingly. Using the default training settings for this dataset resulted in a total training time of 5 hours 3 minutes (0.2 seconds per iteration, for 100,000)). Even though this is a fast and loose comparison (with smaller data), it does suggest that VDSM training time (both stages included) may considerably faster than that of MoCoGAN.

\onecolumn
\centering

\textbf{Generative model}:
\begin{equation}
\begin{split}
    p_{\theta}(\mathbf{x}^n_{t=1:T_n}, \mathbf{s}^n, \mathbf{d}^n, \mathbf{z}^n_{t=1:T_n}) =  p_{\theta}(\mathbf{s}^n)p_{\theta}(\mathbf{d}^n)p_{\theta}(\mathbf{z}_{t=1}^n)\prod_{t=2}^{T_n} p_{\theta}(\mathbf{x}^n_t|\mathbf{s}^n, \mathbf{z}_t^n) p_{\theta}(\mathbf{z}_{t}^n|\mathbf{z}_{t-1}^n, \mathbf{d}^n)
        \end{split}
    \label{eq:genmod}
\end{equation}
\textbf{Inference Model:}
\begin{equation}
\begin{split}
    q_{\phi}(\mathbf{s}^n, \mathbf{d}^n, \mathbf{z}_{t=1:T_n} | \mathbf{x}^n_{t=1:T_n}) = q_{\phi}(\mathbf{s}^n|\mathbf{x}^n_{t=1:T_n})q_{\phi}( \mathbf{d}^n|\mathbf{x}^n_{t=1:T_n})q_{\phi}(\mathbf{z}^n_{t=1},|\mathbf{x}^n_{t=1:T_n}, \mathbf{d}^n) \prod_{t=2}^{T_n}q_{\phi}(\mathbf{z}^n_t| \mathbf{z}^n_{t-1},\mathbf{x}^n_{t=1:T_n}, \mathbf{d}^n)\\
        \end{split}
    \label{eq:infmod}
\end{equation}    
\textbf{ELBO Objective:}
\begin{equation}
    \mbox{max}_{\phi, \theta} \Braket{ \Braket{ \frac{p_{\theta}(\mathbf{x}^n_{t=1:T_n}, \mathbf{s}^n, \mathbf{d}^n, \mathbf{z}^n_{t=1:T_n})}{q_{\phi}(\mathbf{s}^n, \mathbf{d}^n, \mathbf{z}_{t=1:T_n}^n\mid \mathbf{x}^n_{i=1:T_n})) }}_{q_{\phi}}}_{p_{\mathcal{D}(\mathbf{x}^n)}}
    \label{eq:elboreduced}
\end{equation}
\textbf{ELBO Objective (expanded):}
\begin{equation}
\begin{split}
\mathcal{L}(\mathbf{x}^n_{1:T_n};(\theta,\phi)) =\\ \sum_{t=1}^{T_n} \mathbb{E}_{q_{\phi}(\mathbf{z}_t^n, \mathbf{s}^n | \mathbf{x}^n_{1:T_n})}[\log p_{\theta}(\mathbf{x}^n_t|\mathbf{z}_t^n, \mathbf{s}^n)]
-\lambda_{d}(\mbox{KL}(q_{\phi}(\mathbf{d}^n|\mathbf{x}^n_{1:T_n})||p_{\theta}(\mathbf{d}^n))) 
-\lambda_{s}(\mbox{KL}(q_{\phi}(\mathbf{s}^n|\mathbf{x}^n_{1:T_n})||p_{\theta}(\mathbf{s}^n))) \\
-\lambda_{z}(\mbox{KL}(q_{\phi}(\mathbf{z}^n_1|\mathbf{x}^n_{1:T_n}, \mathbf{d}^n)||p_{\theta}(\mathbf{z}^n_1)) 
- \lambda_{z} \sum_{t=2}^{T_n}\mathbb{E}_{q_{\phi}(\mathbf{z}_{t-1}^n|\mathbf{x}^n_{1:T_n},\mathbf{d}^n)}\mbox{KL}(q_{\phi}(\mathbf{z}_t^n|\mathbf{z}_{t-1}^n,\mathbf{d}^n, \mathbf{x}^n_{1:T_n})||p_{\theta}(\mathbf{z}_t^n|\mathbf{z}_{t-1}^n, \mathbf{d}^n)))
\label{eq:full}
\end{split}
\end{equation}
\textbf{Time Varying KL Term (Compact):}
\begin{equation}
\mbox{KL}(q_{\phi}(\mathbf{z}^n_1, ..., \mathbf{z}^n_{T_n}|\mathbf{x}_{1:T_n}, \mathbf{d}^n) ||p_{\theta}(\mathbf{z}_1^n, ...,\mathbf{z}_{T_n}^n| \mathbf{d}^n)))
\label{eq:reduced}
\end{equation}
\textbf{Time Varying KL Derivation:}
\begin{equation}
\begin{split}
\mbox{KL}(q_{\phi}(\mathbf{z}^n_1, ..., \mathbf{z}^n_{T_n}|\mathbf{x}_{1:T_n}, \mathbf{d}^n) ||p_{\theta}(\mathbf{z}_1^n, ...,\mathbf{z}_{T_n}^n| \mathbf{d}^n))) =\\
\int_{\mathbf{z}_1^n}..\int_{\mathbf{z}_{T_n}^n} q_{\phi}(\mathbf{z}^n_1|\mathbf{x}_{1:T_n}, \mathbf{d}^n)...q_{\phi}(\mathbf{z}^n_{T_n}|\mathbf{x}_{1:T_n}, \mathbf{d}^n, \mathbf{z}^n_{T_n-1})
\log\frac{p_{\theta}(\mathbf{z}_1^n, ...,\mathbf{z}_{T_n}^n| \mathbf{d}^n)}{q_{\phi}(\mathbf{z}^n_1|\mathbf{x}_{1:T_n}, \mathbf{d}^n)...q_{\phi}(\mathbf{z}^n_{T_n}|\mathbf{x}_{1:T_n}, \mathbf{d}^n, \mathbf{z}^n_{T_n-1})} =\\
\int_{\mathbf{z}_1^n}..\int_{\mathbf{z}_{T_n}^n}q_{\phi}(\mathbf{z}^n_1|\mathbf{x}_{1:T_n}, \mathbf{d}^n)...q_{\phi}(\mathbf{z}^n_{T_n}|\mathbf{x}_{1:T_n}, \mathbf{d}^n, \mathbf{z}^n_{T_n-1})
\log\frac{p_{\theta}(\mathbf{z}_1^n)p_{\theta}(\mathbf{z}_2^n|\mathbf{z}_1^n, \mathbf{d}^n)...p_{\theta}(\mathbf{z}_{T_n}^n|\mathbf{z}_{T_n-1}^n, \mathbf{d}^n)}{q_{\phi}(\mathbf{z}^n_1|\mathbf{x}_{1:T_n}, \mathbf{d}^n)...q_{\phi}(\mathbf{z}^n_{T_n}|\mathbf{x}_{1:T_n}, \mathbf{d}^n, \mathbf{z}^n_{T_n-1})}=\\
\int_{\mathbf{z}_1^n}..\int_{\mathbf{z}_{T_n}^n}q_{\phi}(\mathbf{z}^n_1|\mathbf{x}_{1:T_n}, \mathbf{d}^n)...q_{\phi}(\mathbf{z}^n_{T_n}|\mathbf{x}_{1:T_n}, \mathbf{d}^n, \mathbf{z}^n_{T_n-1})
\log\frac{p_{\theta}(\mathbf{z}_1^n)}{q_{\phi}(\mathbf{z}_1^n|\mathbf{x}_{1:T_n}^n, \mathbf{d}^n)} \\
+\sum_{t=2}^{T_n}\int_{\mathbf{z}_1^n}..\int_{\mathbf{z}_{T_n}^n}q_{\phi}(\mathbf{z}^n_1|\mathbf{x}_{1:T_n}, \mathbf{d}^n)
...q_{\phi}(\mathbf{z}^n_{T_n}|\mathbf{x}_{1:T_n}, \mathbf{d}^n, \mathbf{z}^n_{T_n-1})\log\frac{p_{\theta}(\mathbf{z}_t^n|\mathbf{z}_{t-1}^n, \mathbf{d}^n)}{q_{\phi}(\mathbf{z}_t^n|\mathbf{z}_{t-1}^n, \mathbf{x}_{1:T_n}^n, \mathbf{d}^n)} =\\
\int_{\mathbf{z}^n_1}q_{\phi}(\mathbf{z}_1|\mathbf{x}_{1:T_n}^n, \mathbf{d}^n)\log\frac{p_{\theta}(\mathbf{z}_1^n)}{q_{\phi}(\mathbf{z}_1^n|\mathbf{x}_{1:T_n}^n, \mathbf{d}^n)} + \sum_{t=2}^{T_n}\int_{\mathbf{z}^n_{t-1}}\int_{\mathbf{z}^n_{t}}q_{\phi}(\mathbf{z}_t^n|\mathbf{x}_{1:T_n}^n, \mathbf{z}^n_{t-1}\mathbf{d}^n)\log\frac{p_{\theta}(\mathbf{z}_t^n| \mathbf{z}_{t-1}^n, \mathbf{d}^n)}{q_{\phi}(\mathbf{z}_t^n|\mathbf{z}_{t-1}^n, \mathbf{x}_{1:T_n}^n, \mathbf{d}^n)}=\\
\mbox{KL}(q_{\phi}(\mathbf{z}^n_1| \mathbf{x}_{1:T_n}^n, \mathbf{d}^n)||p_{\theta}(\mathbf{z}_1^n)) +\sum_{t=2}^{T_n}\mathbb{E}_{q_{\phi}(\mathbf{z}_{t-1}^n|\mathbf{x}^n_{1:T_n},\mathbf{d}^n)} \mbox{KL}(q_{\phi}(\mathbf{z}_t^n|\mathbf{z}_{t-1}^n,\mathbf{d}^n, \mathbf{x}^n_{1:T_n})||p_{\theta}(\mathbf{z}_t^n|\mathbf{z}_{t-1}^n, \mathbf{d}^n)))
\end{split}
\end{equation}
\newpage

\twocolumn

\end{document}